\pdfoutput=1
\documentclass[11pt]{article}

\usepackage[]{acl}

\usepackage{times}
\usepackage{latexsym}
\usepackage{graphicx}
\usepackage{arydshln}
\usepackage{subfigure}
\usepackage{makecell}
\usepackage{colortbl}
\usepackage{pifont}
\usepackage{amssymb}
\usepackage{algorithm}
\usepackage{algorithmic}
\usepackage{enumerate}

\usepackage[T1]{fontenc}
\usepackage{xcolor}
\usepackage[utf8]{inputenc}

\usepackage{microtype}

\usepackage{array}
\usepackage{pifont}
\usepackage{tabularx}
\usepackage{adjustbox}
\usepackage{multirow}
\usepackage{enumitem}
\usepackage{xspace}
\usepackage{tcolorbox}
\usepackage{xparse}
\usepackage{soul}
\usepackage{booktabs,amsfonts,dcolumn}
\usepackage{hyperref}
\usepackage{url}
\usepackage{amsmath,amsthm,amsfonts,amssymb,bm,stmaryrd}
\usepackage[noorphans,vskip=0.75ex,leftmargin=2ex]{quoting}
\usepackage{breqn}
\usepackage{makecell}

\definecolor{OliveGreen}{rgb}{0,0.4,0}

\usepackage{cleveref}

\crefname{section}{§}{§§}
\Crefname{section}{§}{§§}

\newcommand{\ccmark}{\ding{51}}%
\newcommand{\xxmark}{\ding{55}}%
\definecolor{color_m}{RGB}{72,117,170}
\definecolor{color_f}{RGB}{201,89,72}
\definecolor{color_c}{RGB}{230,230,230}
\definecolor{color_e}{RGB}{100,155,74}
\definecolor{ccon}{HTML}{fee9d4}
\definecolor{cood}{HTML}{d8f0d3}
\definecolor{cid}{HTML}{dae8f5}
\definecolor{gg}{HTML}{e2f0cb}

\title{Weight-Inherited Distillation for Task-Agnostic BERT Compression}

\author{
Taiqiang Wu$^{1,2}$\thanks{\ \ Equal contributions. Work was done when Taiqiang and Cheng were interning at Tencent.}, \ Cheng Hou$^{2}$\footnotemark[1],  \
Shanshan Lao$^{3}$,\\
\textbf{Jiayi Li}$^{3}$, \ \textbf{Ngai Wong}$^{1}$, \ \textbf{Zhe Zhao}$^{2}$\footnotemark[2] \and \textbf{Yujiu Yang}$^{3}$\thanks{ \ \ Yujiu Yang and Zhe Zhao are the corresponding authors.}\\
$^{1}$The University of Hong Kong, 
$^{2}$Tencent AI Lab,  \\
$^{3}$ Shenzhen International Graduate School, Tsinghua University \\
\texttt{takiwu@connect.hku.hk}
}

\begin{document}
\maketitle

\begin{abstract}
% The growing size of pre-trained language models~(e.g., BERT and its variants) has led to increased attention in model compression.
% Knowledge Distillation~(KD), which transfers the knowledge from a larger teacher model to a compact student model, is a predominant approach for BERT compression.
% Previous KD-based methods focus on designing extra alignment losses to minimize the distance of logit distributions or intermediate features between teacher and student.
% Previous KD-based methods focus on designing extra alignment losses to minimize the distance of logit distributions or intermediate features between teacher and student.
Knowledge Distillation~(KD) is a predominant approach for BERT compression.
Previous KD-based methods focus on designing extra alignment losses for the student model to mimic the behavior of the teacher model.
These methods transfer the knowledge in an indirect way.
In this paper, we propose a novel Weight-Inherited Distillation~(WID), which directly transfers knowledge from the teacher.
WID does not require any additional alignment loss and trains a compact student by inheriting the weights, showing a new perspective of knowledge distillation.
Specifically, we design the row compactors and column compactors as mappings and then compress the weights via structural re-parameterization.
Experimental results on the GLUE and SQuAD benchmarks show that WID outperforms previous state-of-the-art KD-based baselines.
Further analysis indicates that WID can also learn the attention patterns from the teacher model without any alignment loss on attention distributions.
The code is available at \href{https://github.com/wutaiqiang/WID-NAACL2024}{GitHub}.
% Our key insight is to directly leverage the rich semantic knowledge contained in the weights.
% WID outperforms the previous state-of-the-art KD-based baselines on the GLUE and SQuAD benchmarks, demonstrating the effectiveness of weight inheritance. 
% In particular, it retains 98.9\% and 90.9\% performance of the teacher model adopting only 49.2\% and 10.2\% parameters, respectively.
\end{abstract}

\section{Introduction}

%necessary for bert compression
Transformer-based Pre-trained Language Models~(PLMs), such as BERT \citep{DBLP:conf/naacl/DevlinCLT19}, RoBERTa \citep{DBLP:journals/corr/abs-1907-11692}, XLNET \citep{DBLP:conf/nips/YangDYCSL19}, have achieved great success in many Natural Language Process~(NLP) tasks.
These models are pre-trained on massive corpus via self-supervised tasks to learn contextualized text representations.
However, PLMs have high costs in terms of storage, memory, and computation time, which brings challenges to online services in real-life applications.
Therefore, it is crucial and feasible to compress PLMs while maintaining their performance.

% \begin{table}[!t]
% \centering
% %\tableindent 
% \resizebox{\columnwidth}{!}
% {%
%     \begin{tabular}{lccc}
%         \toprule
%         {\bf Approach} & {\bf Logit Loss } & {\bf Feature Loss} & {\bf Task-Agnostic} \\
%         \midrule
%         DistillBERT & \ccmark & \xxmark & \ccmark \\
%         TinyBERT & \ccmark & \ccmark & \xxmark \\
%         BERT-theseus & \xxmark &\xxmark & \xxmark \\
%         MiniLM & \xxmark & \ccmark & \ccmark \\
%         MobileBERT & \ccmark & \ccmark & \ccmark \\
%         \cellcolor{gg}{WID~(ours)} & \cellcolor{gg}{\xxmark} & \cellcolor{gg}{\xxmark} & \cellcolor{gg}{\ccmark} \\
%         \bottomrule
%     \end{tabular}
% }
% \caption{
% Comparison with previous state-of-the-art distillation methods.
% \textbf{Logit Loss} and \textbf{Feature Loss} denote whether logit-based loss and feature-based loss are used for distillation.
% }
% \label{tab:compare_method}
% \end{table}

\begin{table}[!t]
\centering
%\tableindent 
\resizebox{\columnwidth}{!}
{%
    \begin{tabular}{lcccc}
        \toprule
        \multirow{2}{*}{\bf Approach} & \multicolumn{2}{c}{\bf Alignment Loss} & \multirow{2}{*}{\bf Hard Loss} & \multirow{2}{*}{\bf Task-Agnostic} \\
        \cline{2-3}
        & {\bf Logit } & {\bf Feature} &  & \\
        \midrule
        DistilBERT & \ccmark & \ccmark & \ccmark & \ccmark \\
        TinyBERT~(GD) & \ccmark & \ccmark & \xxmark &  \ccmark \\
        % BERT-of-Theseus & \xxmark &\xxmark & \ccmark &  \xxmark \\
        PKD & \ccmark & \ccmark & \ccmark & \xxmark \\
        MiniLM & \xxmark & \ccmark & \xxmark &  \ccmark \\
        MobileBERT & \ccmark & \ccmark & \ccmark &  \ccmark \\
        \cellcolor{gg}{WID~(ours)} & \cellcolor{gg}{\xxmark} & \cellcolor{gg}{\xxmark} & \cellcolor{gg}{\ccmark}  &  \cellcolor{gg}{\ccmark} \\
        \bottomrule
    \end{tabular}
}
\caption{
Comparison with previous state-of-the-art distillation methods.
\textbf{Logit} and \textbf{Feature} denote whether logit-based loss and feature-based loss are used for distillation.
To the best of our knowledge, WID is the first distillation method without any alignment loss and directly transfers the knowledge by weight inheritance.
}
\label{tab:compare_method}
\end{table}

Knowledge Distillation~(KD), which trains a compact student model by mimicking the behavior of a teacher model, is a predominant method for PLM compression.
% transfer the knowledge in a large teacher model to a compact student model by minimizing their distance.
There are two settings for KD in BERT compression: 1) task-specific, which first fine-tunes the teacher PLMs on specific tasks and then performs distillation, and 2) task-agnostic, which distills PLMs in the pre-training stage.
For task-agnostic distillation, the student model can be directly and generically fine-tuned on various downstream tasks \citep{DBLP:conf/nips/WangW0B0020, DBLP:conf/acl/SunYSLYZ20}.
% Meanwhile, task-agnostic distillation is more challenging.
Hence, we evaluate the proposed weight-inherited distillation~(WID) under a task-agnostic setting.

% problem: with loss
Previous KD-based methods mainly focus on designing alignment losses to minimize the distance between the teacher model and the student model.
We can further categorize these alignment losses into 1) logit-based, which measures the distance of logit distributions, and 2) feature-based, which aims to align the intermediate features including token embeddings, hidden states, and self-attention distributions.
% The logit-based loss measures the distance of logit distributions. 
% % by Kullbak-Leibler~(KL) divergence or Mean Squared Error~(MSE).
% The feature-based loss aims to align the intermediate features including token embeddings, hidden states and self-attention distributions.
% However, adopting these alignment losses brings the following drawbacks:
% 1) selecting various loss functions and balancing the weights of each loss are laborious \citep{DBLP:conf/emnlp/SunCGL19, DBLP:conf/emnlp/JiaoYSJCL0L20};
% 2) some losses will restrict the architecture of the student model.
% For example, attention-based loss \cite{DBLP:conf/emnlp/JiaoYSJCL0L20, DBLP:conf/nips/WangW0B0020, DBLP:conf/acl/SunYSLYZ20} requires the student model to have the same attention heads as the teacher.
% 3) these losses are highly related to transformer structure, thus lacking the potential to compress other models.
However, selecting various loss functions and balancing the weights of each loss are laborious \citep{DBLP:conf/emnlp/SunCGL19, DBLP:conf/emnlp/JiaoYSJCL0L20}.
Meanwhile, the knowledge is embedded in the weights.
This gives rise to an intuitive thought: {\it can we distill the knowledge by directly inheriting the weights, rather than aligning the logit distributions or intermediate features}?

% ours
In this work, we propose Weight-Inherited Distillation~(WID), which does not require any additional alignment loss and trains the student by directly inheriting the weights from the teacher.
In WID, we factorize the KD process into the compression of each weight matrix.
Inspired by structural re-parameterization in CNN compression \citep{DBLP:conf/iccv/DingHT0HGD21}, we design row compactors and column compactors, and then view them as mappings to compress the weights by row and column, respectively.
For the matrices to compress the row only, such as the output layer for MLM task~(the column is always the size of vocabulary), we employ the row compactors exclusively to compress them.
Moreover, during training, we design a novel alignment strategy to align the compactors due to the residual connection in Transformer \citep{DBLP:conf/nips/VaswaniSPUJGKP17}.
% Specifically, 
% Figure \ref{framework} shows the process of compressing a linear layer by WID.
% All compactors are initialized as identity matrices, thus the re-parameterized teacher model produces identical outputs as the original teacher.
% We train the re-parameterized teacher model on the pre-training task and add weight penalty to compactors simultaneously.
% During training, we design the compactor alignment strategy  
% After training, we compress the compactors to desired sizes and merge these compactors and original weights into compact one.
As shown in Table \ref{tab:compare_method}, WID is the only method for task-agnostic distillation without any alignment loss.

% Experiment
We conduct extensive experiments on downstream NLP tasks, including the GLUE and SQuAD benchmarks.
Experimental results demonstrate that WID outperforms traditional KD-based baselines.
%Specifically, the 6-layer model of 768 hidden dimensions distilled from BERTBASE is 2.0× faster, while retaining more than 99\% accuracy on SQuAD 2.0 and several GLUE benchmark tasks.
Further analysis shows that WID can also learn high-level semantic knowledge such as self-attention patterns via inheriting weights.

Our contributions can be summarized as follows:
\begin{itemize}
    % \item We propose a novel Weight-Inherited Distillation~(WID).
    % In WID, the student is learned by inheriting the weight of teacher model without extra alignment loss.
    \item We propose Weight-Inherited Distillation~(WID), revealing a new pathway to KD by directly inheriting the weights via structural re-parameterization.
    \item We design the compactor alignment strategy and conduct WID for task-agnostic BERT compression.
    Experiments on the GLUE and SQuAD benchmark datasets demonstrate the effectiveness of WID for model compression.
    \item We perform further analyses on how to get better performance in BERT compression.
    Even more, we find that WID is able to learn attention patterns from the teacher.
\end{itemize}
\section{Preliminaries}
\label{prelimi}

% Multi-layer Transformers \citep{DBLP:conf/nips/VaswaniSPUJGKP17} have been widely adopted as the backbone of state-of-the-art PLMs.
% In this section, we present a brief introduction to the Transformer and existing KD-based methods.

\subsection{Embedding Layer}
In BERT \citep{DBLP:conf/naacl/DevlinCLT19}, the input texts are tokenized to tokens by WordPiece \citep{DBLP:journals/corr/WuSCLNMKCGMKSJL16}.
% After that, a special classification token~(\texttt{[CLS]}) is added as the first token of each sequence.
The representations~($\{\mathbf{x}_i\}^{|x|}_{i=1}$) of the input sequence are constructed by summing the corresponding token embedding, segment embedding, and position embedding.
% For these three embedding layers in BERT, the weights are $W_{T} \in \mathbb{R}^{|V| \times d}$, $W_{S} \in \mathbb{R}^{2 \times d}$, and $W_{P} \in \mathbb{R}^{512 \times d}$, respectively. 
For the token embedding layer in BERT, the weight is $W_{T} \in \mathbb{R}^{|V| \times d}$, where $|V|$ and $d$ denote the sizes of the vocabulary and the hidden state vector, respectively.

\subsection{Transformer Layer}
Transformer layers are adapted to encode the contextual information of input texts.
The input vector~($\{\mathbf{x}_i\}^{|x|}_{i=1}$) are packed to $\mathbf{H}^0 = [\mathbf{x}_1, \cdots,\mathbf{x}_{|x|} ]$.
After that, the $L$-layer transformer computes the encoding vectors following:
\begin{equation}
    \mathbf{H}^l = \text{Transformer}_{l}(\mathbf{H}^{l-1}), \ l \in [1,L].
\end{equation}
The final output $\mathbf{H}^L = [h^L_1,\cdots,h^L_{|x|}] \in \mathbb{R}^{|x| \times d}$ is employed as the contextualized representation of $\{\mathbf{x}_i\}^{|x|}_{i=1}$.
Each transformer layer consists of a multi-head self-attention~(MHA) sub-layer and a feed-forward~(FFN) sub-layer.
In these two sub-layers, the residual connection \citep{DBLP:conf/cvpr/HeZRS16} is employed, followed by Layer Normalization~(LN)  \citep{DBLP:journals/corr/BaKH16}.

\paragraph{MHA} For the $l$-th transformer layer with $A$ attention heads, the output $\mathbf{O}_{l,a}$ of the attention head $a \in [1,A]$ is calculated as:

\begin{equation}
\begin{aligned}
    \mathbf{Q}_{l,a} &=\mathbf{H}^{l-1}\mathbf{W}^{Q}_{l,a}
    \\ 
    \mathbf{K}_{l,a} &=\mathbf{H}^{l-1}\mathbf{W}^{K}_{l,a}
    \\
    \mathbf{V}_{l,a} &=\mathbf{H}^{l-1}\mathbf{W}^{V}_{l,a}
\end{aligned}
\end{equation}
\begin{equation}
    \mathbf{O}_{l,a} = \mathbf{A}_{l,a} \mathbf{V}_{l,a}, \mathbf{A}_{l,a} =\text{softmax}(\frac{\mathbf{Q}_{l,a} \mathbf{K}^T_{l,a}}{\sqrt{d_k}})
\end{equation}
where linear projection $\mathbf{W}^{Q}_{l,a},\mathbf{W}^{K}_{l,a},\mathbf{W}^{V}_{l,a} \in \mathbb{R}^{d \times d_k}$ and $d_k=\frac{d}{A}$ is the dimension of each head.
The final output of MHA sub-layer is as follows:
\begin{equation}
    \mathbf{O}_{l} = \text{LN}(\mathbf{H}^{l-1} + (||^{A}_{a=1}\mathbf{O}_{l,a}) \mathbf{W}^{O}_{l})
\end{equation}
where $\mathbf{W}^{O}_{l} \in \mathbb{R}^{d \times d}$, $\text{LN}$ is layer normalization and $||$ denotes the concatenation operation.
%$\mathbf{b}^{O}_{l} \in \mathbb{R}^{d}$,

\paragraph{FFN} The $l$-th FFN sub-layer consists of an up projection and a down projection, parameterized by $\mathbf{W}^{U}_{l} \in \mathbb{R}^{d \times d_f}$, $\mathbf{W}^{D}_{l} \in \mathbb{R}^{d_f \times d}$, and corresponding bias $\mathbf{b}^{U}_{l} \in \mathbb{R}^{d_f}$, $\mathbf{b}^{D}_{l} \in \mathbb{R}^{d}$:
\begin{equation}
    \text{FFN}(\mathbf{O}_{l}) = \text{gelu}(\mathbf{O}_{l} \mathbf{W}^{U}_{l} + \mathbf{b}^{u}_{l}) \mathbf{W}^{D}_{l} + \mathbf{b}^{d}_{l}.
\end{equation}
Typically, $d_f=4d$.
Finally, we obtain the output of layer $l$ by:
\begin{equation}
    \mathbf{H}^{l} = \text{LN}(\mathbf{O}_{l} + \text{FFN}(\mathbf{O}_{l})).
\end{equation}

\subsection{Knowledge Distillation}
% Knowledge Distillation~(KD) aims to transfer the knowledge from teacher model $T$ to compact student model $S$.
% The student model $S$ is trained to mimic the behaviors of teacher model $T$ via minimizing the distance between them.
Knowledge Distillation~(KD) trains a compact student model $S$ by mimicking the behaviors of the teacher model $T$.
The losses can be categorized into logit-based and feature-based.

For logit-based loss, the target is to minimize the distance between logit distribution $\mathbf{p}_s$ from the student and $\mathbf{p}_t$ from the teacher, which can be formalized as:
\begin{equation}
    \mathcal{L}_{logit} = \mathcal{H}_1(\mathbf{p}_s/\tau,\mathbf{p}_t/\tau),
\end{equation}
where $\tau$ is the temperature and $\mathcal{H}_1$ is the cross-entropy loss or KL-divergence.

Feature-based loss aims to align the intermediate features between the teacher and the student by:
\begin{equation}
    \mathcal{L}_{feature} = \mathcal{H}_2 (f^S(x), f^T(x)),
\end{equation}
where $\mathcal{H}_2$ is the loss function such as Mean Square Error~(MSE) and $f(x)$ denotes for the intermediate output including hidden state vector $\mathbf{H}$ and attention distribution $\mathbf{A}$.

As shown in Table \ref{tab:compare_method}, logit-based and feature-based loss can be jointly employed for better distillation.
However,  balancing the weights of each loss is laborious.
For example, the overall loss of PKD \citep{DBLP:conf/emnlp/SunCGL19} is:
\begin{equation}
    \mathcal{L} = (1-\alpha) \mathcal{L}_{hard} + \alpha \mathcal{L}_{logit} + \beta \mathcal{L}_{feature},
\end{equation}
where $\mathcal{L}_{hard}$ is the loss on target tasks and $\alpha$ and $\beta$ are the hyper-parameters.
PKD performs grid search over $\alpha$ and $\tau$, where $\alpha \in \{0.2,0.5,0.7\}$ and $\tau \in \{5,10,20\}$.
After that, the best $\alpha$ and $\tau$ are fixed, followed by a search of $\beta \in \{10,100,500,1000\}$.

Meanwhile, selecting various loss functions is also laborious.
In PKD, $\mathcal{L}_{feature}$ is defined as the mean square loss
between the normalized hidden states for each layer.
DistilBERT \citep{DBLP:journals/corr/abs-1910-01108} adopts the cosine embedding loss for hidden states.
TinyBERT \citep{DBLP:conf/emnlp/JiaoYSJCL0L20} employs the mean square loss for self-attention distributions, embedding layer outputs, and hidden states.
\section{Weight-Inherited Distillation}

% Here we introduce the details of WID and the compactor alignment strategy.
% In this section, we propose a novel Weight-Inherited Distillation~(WID) method for transformer-based models without any alignment loss. 
% The WID aims to directly leverage knowledge in weight and compress the teacher model by learning mappings for the compact student model.

\begin{figure}[t]
	\centering
	\includegraphics[width=0.98\linewidth]{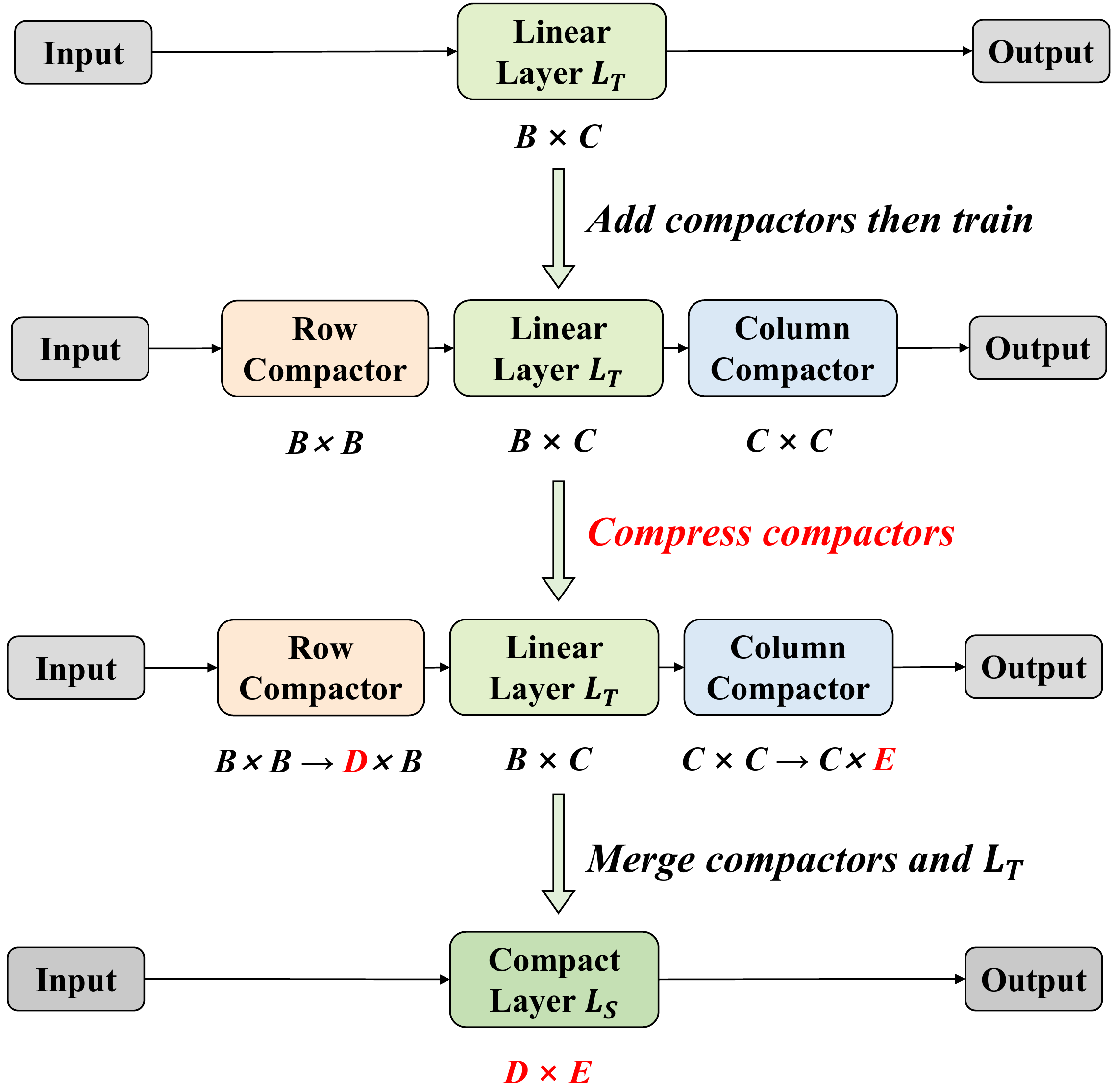}
	\caption{
	Overview of compressing linear layer $L_T$ with weight $\mathbf{W}^{L_T} \in \mathbb{R}^{B \times C}$ to compact linear layer $L_S$ with weight $\mathbf{W}^{L_S} \in \mathbb{R}^{D \times E}$ via WID.
	Both row compactor and column compactor are initialized as {\bf identity matrices}.
	After training, we compress the compactors and merge them with the original layer.
	All the linear layers in the teacher model are compressed {\bf simultaneously}.
% The \underline{input} for this layer is changed since the previous layer is compressed simultaneously.
% 	For embedding layers, we only adopt the column compactor since the row size is not changed in their weights.
    }
	\label{framework}
	% \vspace{-0.5em}
\end{figure}

\subsection{Structural Re-parameterization}

As mentioned in Section \ref{prelimi}, the PLMs~(e.g., BERT) consist of embedding layers and transformer layers.
% Among these layers, we can divide all the operations into linear layer and non-linear activation function.
To compress the BERT, we have to learn a mapping from the larger weight in the teacher model to the compact one.
% In WID, we adopt the structural re-parameterization~(\citep{DBLP:conf/iccv/DingHT0HGD21}) to compress the weights. 
% Since some 
% we design the row compactors and column compactors.
In terms of matrices, these mappings can be categorized as: 
% 1)  {\bf column mapping}, such as the token embedding matrix $\mathbf{W}_T \in \mathbb{R}^{|V| \times d}$, 2)  {\bf row mapping}, such as the output layer for MLM with size $\mathbb{R}^{d \times |V|}$, and 3) {\bf column and row mapping}, such as up projection $\mathbf{W}_{l,u} \in \mathbb{R}^{d \times d_f}$ in FFN.
\begin{itemize}
    \item column mapping only, such as the token embedding matrix $W_T \in \mathbb{R}^{|V| \times d}$,
    \item row mapping only, such as the weight of output layer for MLM task with size $\mathbb{R}^{d \times |V|}$,
    \item column and row mapping, such as up projection $\mathbf{W}_{l,u} \in \mathbb{R}^{d \times d_f}$ in FFN.
\end{itemize}
In WID, we adopt the re-parameterization trick and design the row compactor for row mapping and column compactor for column mapping, respectively.
% The column and row mapping is viewed as the compound of row mapping and column mapping, and thus we adopt row compactor and column compactor together.

\begin{figure*}[!t]
    % \vspace{-1em}
	\centering
	\includegraphics[width=\linewidth]{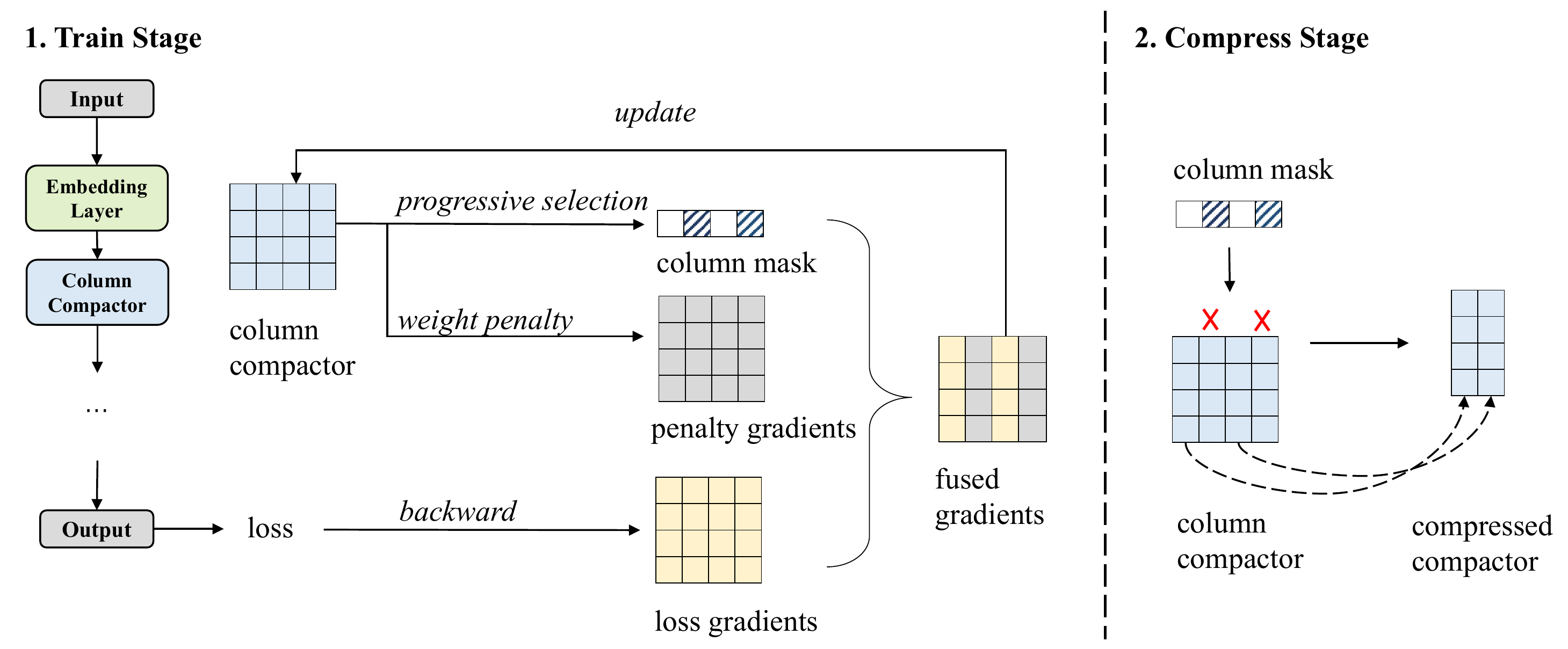}
	\caption{
	Training and compression for column compactor.
	During the training process, we add weight penalty gradients by columns and progressively select the mask to fuse the penalty gradients and original loss gradients.
% 	For gradients fusion, we decouple penalty gradients and original loss gradients to avoid gradient competition.
	After training, we compress the column compactor following the column mask.
	}
	\label{column_compress}
\end{figure*}

Figure \ref{framework} gives an example showing the process of compressing the original weight $\mathbf{W}^{L_T} \in \mathbb{R}^{B \times C}$ to a compact weight $\mathbf{W}^{L_S} \in \mathbb{R}^{D \times E}$ adopting both row compactor and column compactor.
First, we insert the row compactor with weight $\mathbf{W}^{rc} \in \mathbb{R}^{B \times B}$ and the column compactor with weight $\mathbf{W}^{cc} \in \mathbb{R}^{C \times C}$ before and after the linear layer $L_T$ from the teacher model.
All compactors are linear layers without bias and their weights are initialized as identity matrices.
For an arbitrary input $X$, the re-parameterized teacher model produces identical outputs as the original, since
\begin{equation}
    X\mathbf{W}^{L_T} = X \mathbf{W}^{rc} \mathbf{W}^{L_T} \mathbf{W}^{cc}.
\end{equation}
Second, we train the re-parameterized teacher model on the pre-training task.
% During training, we add the row/column penalty to some rows/columns of the row/column compactor.
% The norm of these rows/columns with penalty become as small as possible.
% Therefore, these rows/columns become less and less important to the performance. 
After training, the row compactor is compressed by reducing the $B-D$ rows, and the column compactor is compressed by reducing $C-E$ columns.
% The goal is to maintain the performance of the teacher model by the rows/column without penalty and compress the compactor simultaneously.
The objects are as follows:
\begin{equation}
\begin{aligned}
    \mathbf{W}^{rc} \in \mathbb{R}^{B \times B} & \rightarrow \mathbf{W}^{rc^{\prime}} \in \mathbb{R}^{D \times B} \\
    \mathbf{W}^{cc} \in \mathbb{R}^{C \times C} & \rightarrow \mathbf{W}^{cc^{\prime}} \in \mathbb{R}^{C \times E}.
\end{aligned}
\end{equation}
More details can be found in Section \ref{comp-compress}.
Finally, we merge the compressed compactors $\mathbf{W}^{rc^{\prime}}, \mathbf{W}^{cc^{\prime}}$ and the original teacher layer $\mathbf{W}^{L_T}$ to obtain the compact layer for the student following:
\begin{equation}
\label{eq_merge}
    \mathbf{W}^{L_S} = \mathbf{W}^{rc^{\prime}} \mathbf{W}^{L_T} \mathbf{W}^{cc^{\prime}} \in \mathbb{R}^{D \times E}
\end{equation}
% Specially, the input is changed since the previous layer is compressed simultaneously.

% For row mapping and column mapping, we employ the row compactor and column compactor respectively.
% Moreover, we freeze the original linear layer from the teacher model and only update the compactors to avoid over-fitting.
For the weights to compress the rows only, we adopt the row compactor exclusively. 
Similarly, we employ the column compactor exclusively for the weights to compress the columns only.

\subsection{Compactor Compression}
\label{comp-compress}
% In WID, we design row compactors and column compactors and view them as mappings to compress the weights by row and column, respectively.
The goal is to maintain the performance of the teacher model as much as possible and compress the compactors simultaneously.
% Compared to directly learning these compactors, our key insight is to initialize these compactors with identity matrices and compress them to the desired size progressively.

Figure \ref{column_compress} presents the training and compression process for the column compactor.
% To compress the column compactor $W^{CC} \in \mathbb{R}^{C \times C}$, we fuse the train loss gradients and extra penalty gradients to update the 
% we compress the compactor progressively and dynamically.
% During the training process, we updated
To compress the compactors, we add extra penalty loss to minimize the norms of some columns.
Given the column compactor $\mathbf{W}^{cc} \in \mathbb{R}^{C \times C}$ and original gradients $g^{cc}_{ori} \in \mathbb{R}^{C \times C}$ from training tasks, the penalty gradients $g^{cc}_{pen} \in \mathbb{R}^{C \times C}$ are calculated as follows:
\begin{equation}
\label{eq_pe}
    g^{cc}_{pen} = \frac{\mathbf{W}^{cc}}{||\mathbf{W}^{cc}||_2}
\end{equation}
where $||\mathbf{W}^{cc}||_2$ denotes the Euclidean norm across each column.

However, applying the $g^{cc}_{ori}$ and penalty gradients $g^{cc}_{pen}$ to the same row/column leads to the gradient competition~\citep{DBLP:conf/iccv/DingHT0HGD21}.
Therefore, we choose some columns to reduce and apply the penalty gradients $g^{cc}_{pen}$, while the rest columns are adopted to keep performance and updated with $g^{cc}_{ori}$.
Specifically, we pick top-$k$ columns with lower norm value based on the $||\mathbf{W}^{cc}||_2$ and set the corresponding value in our column mask $M =\{0,1\}^C$ to be 1.
Later, the original gradients $g^{cc}_{ori}$ and the penalty gradients $g^{cc}_{pen}$ are fused as follows:
\begin{equation}
\label{eq_fuse}
    g^{cc}_{fused}[:,i] = 
    \begin{cases}
    g^{cc}_{pen}[:,i], \ & \text{if} \ M[i]=1 \\
    g^{cc}_{ori}[:,i], \ & \text{if} \  M[i]=0
    \end{cases}
\end{equation}
where $0 \leq i \leq C$.
We employ the fused gradients $g^{cc}_{fused}$ to update the corresponding column compactor.
After training, we compress the column compactor by column mask:
\begin{equation}
\label{eq_prune}
    \mathbf{W}^{cc^{\prime}} = \mathbf{W}^{cc}[:,i], \ \text{where} \ M[i]=0.
\end{equation}
Moreover, the process is similar for row compactors.
We calculate $||\mathbf{W}^{rc}||_2$ for each row and select the top-$k$ rows with the lower norm value. 

\begin{figure*}[!t]
    \vspace{-1em}
	\centering
	\includegraphics[width=\linewidth]{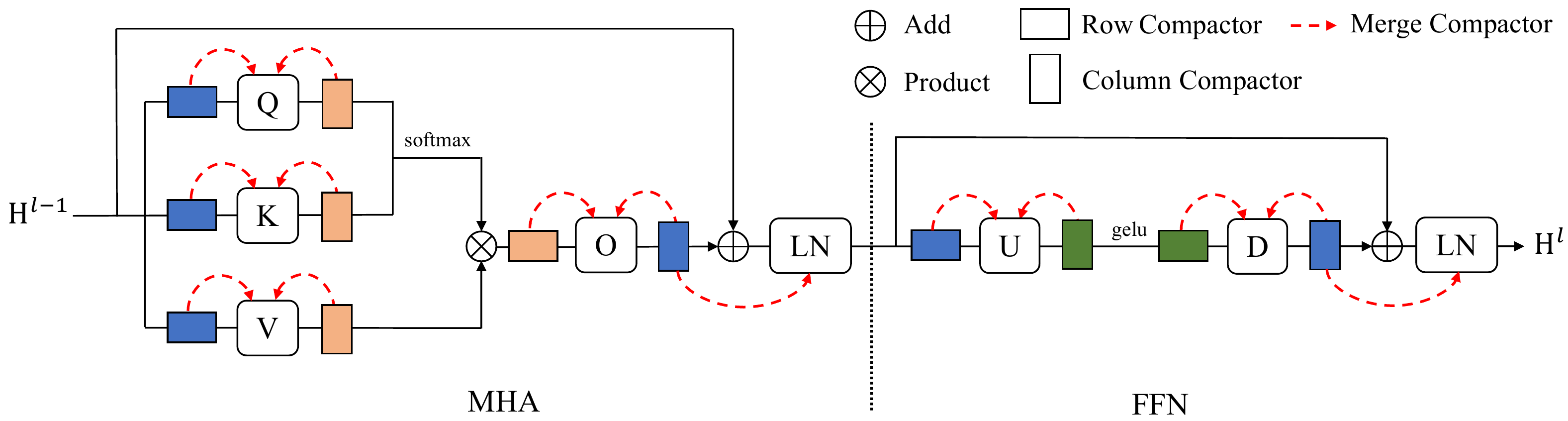}
	\caption{
	Compactor merging process for a Transformer block.
	For the bias terms, we merge them with corresponding column compactors.
	For beta and gamma in Layer Norm~(LN), we adopt the previous column compactors to update them.
	During training, the compactors in the \textbf{same color} are aligned.
	For each group of the aligned compactors, we learn one of them and duplicate~(or, flip) it for the rest compactors.
	}
	\label{align_compactor}
	% \vspace{-0.5em}
\end{figure*}

For stability and better performance, we choose the rows/columns of the compactors progressively.
Concretely, we increase $k$ by $d$ for $N$ steps until reaching the desired size during the training stage.
% The algorithm is shown in Appendix \ref{appendix-alo}.
Moreover, we also try the dynamic selection \citep{DBLP:conf/iccv/DingHT0HGD21} for mask and it makes no effect.

\subsection{Compactor Alignment Strategy}
To apply WID for BERT compression, we design a novel compactor alignment strategy.
Since each dimension in a hidden representation $h_1$ is connected to the same dimension in another hidden representation $h_2$ through a residual connection, the compactors before and after the $h_1$ and $h_2$ need to be aligned.
As shown in Figure \ref{align_compactor}, the compactors in a transformer block are divided into three groups~(same color, same group).
The first compactor before the $\mathbf{H}^{l-1}$ and the first compactor after the $\mathbf{H}^{l}$ are also aligned with groups in blue.
Therefore, the column compactor for the embedding layer, the row compactor for the output layer, and compactors in blue from each layer are all aligned.
Meanwhile, the groups in orange/green can be different across layers since they are not adjacent.
For each group of the aligned compactors, we learn one of them and duplicate~(or, flip) it for the rest. 
Please refer to Appendix \ref{appendix-align} for more details.
% Specifically, we can align all the compactors in BERT into the following groups:
% \begin{itemize}
%     \item group in blue: \{RC for embedding layer, blue compactors in each Transformer layer, LC for output layer\},
%     \item groups in orange: \{orange compactors in layer$1$\}; \{orange compactors in layer2\}; ... \{orange compactors in layer$N$\},
%     \item groups in green:  \{green compactors in layer1\}; \{green compactors in layer2\}; ... \{green compactors in layer$N$\}.
% \end{itemize}

\section{Experiments}
% In this section, we present the empirical results on the GLUE and SQuAD benchmarks between WID and various baselines.

\subsection{Task-Agnostic Distillation}
We employ the uncased version of $\text{BERT}_{\text{base}}$ as our teacher model \footnote{From https://huggingface.co/bert-base-uncased} and implement WID based on TencentPretrain framework\citep{zhao2023tencentpretrain}.
$\text{BERT}_{\text{base}}$ \citep{DBLP:conf/naacl/DevlinCLT19} is a 12-layer transformer model~($d$=768, $A$=12, $L$=12), which contains 110M parameters.
For student models, we compress the teacher model to various model sizes for comparison, including $\text{WID}_{55}$~($d$=516, $A$=12, $L$=12) with 55M parameters and $\text{WID}_{11}$~($d$=192, $A$=12, $L$=12) with 11M parameters.
We use the documents of English Wikipedia and BookCorpus \citep{DBLP:conf/iccv/ZhuKZSUTF15} for pre-training following \citet{DBLP:conf/naacl/DevlinCLT19}.
We use AdamW \citep{DBLP:conf/iclr/LoshchilovH19} with $\beta_1=0.9$, $\beta_2=0.99$.
The compactors are trained with peak learning rate 5e-5 and the original linear layers with peak learning rate 1e-6.
For WID, we adopt the 2-norm and set $N$=$500$, $d$=$\lfloor ({d}_{t}-{d}_{s})/16 \rfloor$.
It costs about 64 hours to train for 400,000 steps with a batch size of 960 on 8 A100 GPUs.

\subsection{Downstream Tasks}
Following previous PLM-based KD methods \citep{DBLP:journals/corr/abs-1910-01108, DBLP:conf/nips/WangW0B0020}, we evaluate our WID on the SQuAD v1.1 \citep{DBLP:conf/emnlp/RajpurkarZLL16} and GLUE benchmark \citep{DBLP:conf/iclr/WangSMHLB19}.
The GLUE benchmark consists of CoLA \citep{DBLP:journals/tacl/WarstadtSB19}, SST-2\citep{DBLP:conf/emnlp/SocherPWCMNP13}, MRPC \citep{DBLP:conf/acl-iwp/DolanB05}, STS-B \citep{DBLP:conf/semeval/CerDALS17}, QQP \citep{chen2018quora}, MNLI \citep{DBLP:conf/naacl/WilliamsNB18}, QNLI\citep{DBLP:conf/emnlp/RajpurkarZLL16} and RTE \citep{DBLP:conf/tac/BentivogliMDDG09}.
After task-agnostic distillation, we fine-tune our compressed BERT WID$_{55}$ and WID$_{11}$ on these benchmarks adopting the grid search and report the results on the development sets.
The result of MNLI is the score of MNLI-m.
% For SQuAD, we report EM and F1.
More details about these datasets including dataset sizes and metrics and the hyperparameters for fine-tuning can be found in the Appendix \ref{app:data}.

\begin{table*}[!t]
\centering
% \tableindent 
\resizebox{2\columnwidth}{!}
{%
    \begin{tabular}{l|cc|ccccccccc|c}
        \toprule
        % \multirow{2}{*}{\bf Method} & \multirow{2}{*}{\bf FLOPS \\ 1e10} & \multirow{2}{*}{\bf Params} & \multirow{2}{*}{\bf SST-2} & \multirow{2}{*}{\bf CoLA} & \multirow{2}{*}{\bf MRPC} & \multirow{2}{*}{\bf QNLI} & \multirow{2}{*}{\bf QQP} & \multirow{2}{*}{\bf RTE} & \multirow{2}{*}{\bf STS-B} & \multirow{2}{*}{\bf MNLI} & \multirow{2}{*}{\bf SQuAD}& \multirow{2}{*}{\bf AVG} \\
        % &  &  &  &  &  &  &  &  &  &  &  &   \\
        {\bf Method} & {\bf FLOPs} & {\bf Params} & {\bf SST-2} & {\bf CoLA} & {\bf MRPC} & {\bf QNLI} & {\bf QQP} & {\bf RTE} & {\bf STS-B} & {\bf MNLI} & {\bf SQuAD} & {\bf AVG} \\
        \midrule
        $\text{BERT}_{\text{base}}$ & 22.7B & 110.1M & 92.7 & 59.1 & 90.4 & 91.7 & 91.4 & 70.8 & 90.1 & 84.5 & 89.6/82.6 & 84.3 \\
        \midrule
        $\text{DistilBERT}$      & 11.9B & 67.5M & 91.3 & 51.3 & 87.5 & 89.2 & 88.5 & 59.9 & 86.9 & 82.2 & 86.2/78.1 & 80.1 \\
        MiniLM                      & 11.9B & 67.5M & 92.0 & 49.2 & 88.4 & 91.0 & 91.0 & 71.5 & - & 84.0 & -/- & - \\
        MiniLM v2                   & 11.9B & 67.5M & 92.4 & 52.5 & 88.9 & 90.8 & 91.1 & 72.1 & - & 84.2 & -/-  & - \\
        TinyBERT~(GD)$^\dagger$     & 11.9B & 67.5M & 92.9 & 44.1 & 89.5 & 90.7 & 91.0 & 73.7 & 89.6 & 83.8 & 84.0/74.2 & 81.3 \\
        TinyBERT~(GD)$^\ddagger$    & 10.4B & 54.9M & 92.3 & 47.0 & 87.3 & 90.8 & 90.9 & 69.7 & 89.0 & 83.3 & 85.4/76.2 & 81.2 \\
        \rowcolor{gray!15} $\text{WID}_{\text{55}}$~(ours)    & 10.4B & 54.9M &  92.4 & 61.7 & 88.2 & 90.1 & 91.0 & 70.4 & 87.9 & 82.9 & 88.5/80.8 & \textbf{83.4} \\
        \midrule
        TinyBERT~(GD)$^\ddagger$    & 1.6B & 11.3M & 88.4 & 30.3 & 80.4 & 87.5 & 89.1 & 65.3 & 84.0 & 79.4 & 80.5/70.7 & 75.6 \\
        \rowcolor{gray!15} $\text{WID}_{\text{11}}$~(ours)    & 1.6B & 11.3M & 88.8 & 44.2 & 81.9 & 85.4 & 89.5 & 60.3 & 84.5 & 78.4 & 81.2/72.4 & \textbf{76.7} \\
        \bottomrule
    \end{tabular}
}
\caption{
Comparison between our WID and various task-agnostic distillation methods.
% For SQuAD v1.1, we report the F1/EM scores.
We compare the task-agnostic distilled models without both data augmentation and task-specific distillation.
% WID achieves better performances than TinyBERT under various model size. 
$\dagger$ means that we fine-tune the official weights.
$\ddagger$ means that we reproduce the methods following the official code.
Other results are taken from corresponding papers.
For MiniLM and MiniLM v2, the average reported scores are 81.0 and 81.7, and both are lower than the 82.3 of WID.
}
\label{tab: main_res}
% \vspace{-1em}
\end{table*}

\subsection{Baselines}
For a fair comparison, we compare our WID with the \textbf{task-agnostic distillation} baselines.
These baselines include: 1) DistilBERT \citep{DBLP:journals/corr/abs-1910-01108}, which distills the student by the combination of the original MLM loss, the cosine distance for features, and the KL divergence for output logits. 2) TinyBERT (GD) \citep{DBLP:conf/emnlp/JiaoYSJCL0L20}, which aligns the attention distributions and hidden states for general distillation.  3) MiniLM \citep{DBLP:conf/nips/WangW0B0020} and MiniLM v2 \citep{DBLP:conf/acl/WangBHDW21}, which align the attention matrix and values-values scaled dot-product.
% 3) BERT-of-Theseus \citep{DBLP:conf/emnlp/XuZGWZ20}, which compresses the teacher model by progressive module replacing.
We also reproduce the TinyBERT in the same architecture as WID, following the official code.
For fair comparison, we employ the same corpus and follow the official hyperparameters.
We do not compare with MobileBERT \citep{DBLP:conf/acl/SunYSLYZ20} since its teacher is IB-BERT$_{\text{large}}$~(much higher accuracy than $\text{BERT}_{\text{base}}$) and its computations~(4096 batch size, 740,000 steps) is much higher.
Moreover, we also compare WID with task-specific methods in Appendix \ref{appen: task-spec}.

\subsection{Main Results}

We compare WID with other task-agnostic distillation methods in \textbf{various} model sizes.
All the methods utilize the $\text{BERT}_{\text{base}}$ as the teacher model.
As shown in Table \ref{tab: main_res}, WID retains 98.9\% and 90.9\% performance of $\text{BERT}_{\text{base}}$ using only 49.2\% and 10.2\% parameters, respectively.
In particular, in the CoLA task, $\text{WID}_{\text{55}}$ gets a higher score than $\text{BERT}_{\text{base}}$. 
Compared to the baselines with 67.5M parameters, $\text{WID}_{\text{55}}$ gets comparable performance with MiniLM and higher performance than DistilBERT with fewer parameters.
Meanwhile, WID outperforms the TinyBERT under the same architecture on  GLUE benchmarks and SQuAD, showing its superiority over the traditional KD methods with logit-based loss and feature-based loss.
Without CoLA, $\text{WID}_{\text{55}}$ gets an average score of 85.8 and still outperforms the TinyBERT~(GD) with an average score of 85.0. 

Meanwhile, we apply WID for generative PLM.
Please refer to \ref{appendix: gpt_compress} for more details.

\paragraph{Larger Performance Gap}
Since the performance gap between teacher and student has always been a crucial point and difficulty in KD, we conduct experiments for smaller student models~(11.3M parameters).
We reproduce the task-agnostic TinyBERT under the General Distillation~(GD) as the baseline.
% For fair comparison, we utilize the same corpus and hyper-parameters.
As shown in Table \ref{tab: main_res}, we find that WID~(average score: 76.7) still outperforms TinyBERT~(average score: 75.6) when the student model is about 10x smaller.
% Experimental results on various student model sizes demonstrate the effectiveness of WID comparing to the traditional KD methods with extra alignment loss. 

\section{Analysis and Discussion}

\begin{table*}[!t]
\centering
% \tableindent 
\resizebox{1.9\columnwidth}{!}
{%
    \begin{tabular}{l|ccccccccc|c}
        \toprule
        {\bf Method} & {\bf SST-2} & {\bf CoLA} & {\bf MRPC} & {\bf QNLI} & {\bf QQP} & {\bf RTE} & {\bf STS-B} & {\bf MNLI} & {\bf SQuAD} & {\bf AVG} \\
        \midrule
        $\text{WID}_{\text{55}}^{dim}$   &  92.4 & 61.7 & 88.2 & 90.1 & 91.0 & 70.4 & 87.9 & 82.9 & 88.5/80.8 & \textbf{83.4} \\
        $\text{WID}_{\text{55}}^{head}$  &  92.0 &  61.6 &  88.2 &  89.4 &  91.0 &  70.8 &  87.6 &  82.6 &  87.3/79.4 & 83.0 \\
        \midrule
        $\text{WID}_{\text{11}}^{dim}$  & 88.8 & 44.2 & 81.9 & 85.4 & 89.5 & 60.3 & 84.5 & 78.4 & 81.2/72.4 & 76.7 \\
        $\text{WID}_{\text{11}}^{head}$ & 89.6 & 46.2 & 83.1 & 86.1 & 89.5 & 62.1 & 85.3 & 79.0 & 81.7/72.9 & \textbf{77.6} \\
        \bottomrule
    \end{tabular}
}
\caption{
Comparison between dropping heads and reducing dimension of each head for $\text{WID}_{\text{55}}$ with 55M parameters and $\text{WID}_{\text{11}}$ with 11M parameters.
}
\label{tab: head_or_dim}
% \vspace{-1em}
\end{table*}
\begin{table*}[!t]
\centering
% \tableindent 
\resizebox{2.0\columnwidth}{!}
{%
    \begin{tabular}{l|c|ccccccccc|c}
        \toprule
        {\bf Teacher} & {\bf Params} & {\bf SST-2} & {\bf CoLA} & {\bf MRPC} & {\bf QNLI} & {\bf QQP} & {\bf RTE} & {\bf STS-B} & {\bf MNLI} & {\bf SQuAD} & {\bf AVG} \\
        \midrule
        $\text{BERT}_{\text{base}}$ & 110.1M & 89.6 & 46.2 & 83.1 & 86.1 & 89.5 & 62.1 & 85.3 & 79.0 & 81.7/72.9 & 77.6 \\
        $\text{BERT}_{\text{55}}$   & 54.2M & 89.5 & 43.2 & 84.6 & 86.3 & 89.7 & 63.2 & 85.7 & 79.4 & 81.2/72.5 & 77.5 \\
        $\text{WID}_{\text{55}}^{head}$ & 54.2M & 89.9 & 46.2 & 84.8 & 86.5 & 89.5 & 64.6 & 84.7 & 78.8 & 82.1/73.5 & \textbf{78.1} \\
        \bottomrule
    \end{tabular}
}
\caption{
Comparison between different teacher models after they are compressed to $\text{WID}_{\text{11}}^{head}$.
$\text{BERT}_{\text{55}}$ means the BERT model with same architecture as $\text{WID}_{\text{55}}^{head}$.
% Both $\text{BERT}_{\text{base}}$ and $\text{BERT}_{\text{55}}$ are downloaded from the Hugging face \footnote{https://huggingface.co/google/bert_uncased_L-12_H-768_A-12}.
% We can find that knowledge is accumulating through more generations’ involvements
}
\label{tab: distill_over_gene}
% \vspace{-0.5em}
\end{table*}

% \subsection{Compare WID with Pruning and Self-Distillation}
\subsection{WID vs Pruning}
% We propose WID, a weight-inherited distillation method for task-agnostic BERT compression without extra alignment loss, which learns mappings from the teacher model to compact student via re-parameterization.
% To compress the linear layer, we design the row compactor and column compactor for row squeezing and column squeezing, respectively.
% These compactors are initialized with identity matrices and trained with weight penalty.
% After training, we prune the compactors and merge them with the original layer to obtain the compact student model.
% We propose WID, a weight-inherited distillation method via structural re-parameterization.
% , which learns mappings from the teacher model to compact student via structural re-parameterization.
% and self-distillation\citep{DBLP:conf/iccv/ZhangSGCBM19}.
Pruning \citep{DBLP:conf/nips/CunDS89} aims to remove redundant weights from a neural network to achieve parameter-efficiency while preserving model performance, including unstructured pruning which sets weights to 0, and structured pruning which removes components such as attention heads.
Unstructured pruning methods do not reduce the model size.
However, WID is very likely to be confused with structured pruning methods.

Structured pruning methods aim to remove the redundant units and then usually get sub-networks without a pre-defined structure.
However, WID \textbf{does not} remove any parts of the original weights from the teacher models but learns a student model with a pre-defined structure.
% the structure of the student model \textbf{is} pre-defined.
Meanwhile, the goal of KD is to transfer the knowledge from teacher models to student models.
In WID, we design the compactors as mappings to inherit knowledge from teacher models, rather than to find sub-networks.
Hence, we consider WID as a KD method though the compression process of compactors is similar to pruning.
More comparison between WID and pruning methods can be found in \ref{appendix: pruning}.
% WID, as a knowledge distillation method, differs from the structured pruning methods as follows: 

% (1)~Structured pruning methods aim to remove the redundant parts and then get sub-networks without pre-defined structure.
% 

% (2)~

% In WID, we do not remove any parts of the weights from the teacher model.
% Instead of that, we adopt the row/column compactors to linearly combine the rows/columns of the weights into smaller size.
% Therefore, the knowledge of teacher models can be transferred to student models.
% via structural re-parameterization.
% Actually, we borrow the idea from pruning to compress the compactors.
% Meanwhile, the structure of the student model is pre-defined in WID while the pruning methods usually do not fix the structure of compact model.
% WID aims to transfer the knowledge from teacher models to student models, rather than find good subnetworks.

% Self-distillation\citep{DBLP:conf/iccv/ZhangSGCBM19} is a one-step {\bf online distillation} method, which distills the knowledge in deeper layer to shallow layer during the training process of teacher model.
% Compared to self-distillation, WID is an {\bf offline distillation} method, since the teacher model is trained before knowledge distillation.
% Furthermore, self-distillation aims to transfer knowledge by aligning intermediate features or logit distributions, while WID transfers knowledge by inheriting the weight directly.

\subsection{MHA: Dropping Heads or Reducing Dimension}
Multi-Head Attention~(MHA) allows the model to jointly attend to the information from different representation subspaces \citep{DBLP:conf/nips/VaswaniSPUJGKP17}.
When compressing the weights in MHA, there are two options, including 1) dropping heads, which reduces the number of heads $A$, and 2) reducing dimension, which reduces the size of each head $d_k$.
For TinyBERT \citep{DBLP:conf/emnlp/JiaoYSJCL0L20} and MiniLM \citep{DBLP:conf/nips/WangW0B0020}, they keep $A$=12 and reduce $d_k$ due to the constraint of attention-based loss.
Our proposed WID is more flexible since we do not employ any alignment loss.
Moreover, we can easily achieve these two strategies by constraining the column mask in MHA.
For $\text{WID}_{\text{55}}$ and $\text{WID}_{\text{11}}$ reported in Table \ref{tab: main_res}, we reduce the size of each attention head following TinyBERT for a fair comparison.

To further explore these two strategies, we conduct WID under these two settings and report the scores on downstream tasks. 
In $\text{BERT}_{\text{base}}$, we have $A$=12 and $d_k$=64.
The student models are selected as: $\text{WID}_{\text{55}}^{dim}$~($A$=12, $d_k$=43), $\text{WID}_{\text{55}}^{head}$~($A$=8, $d_k$=64),
$\text{WID}_{\text{11}}^{dim}$~($A$=12, $d_k$=16),
and $\text{WID}_{\text{11}}^{head}$~($A$=3, $d_k$=64).
As shown in Table \ref{tab: head_or_dim}, the dropping head strategy performs slightly worse under 55M parameters and much better under 11M parameters.
% For $\text{WID}_{\text{55}}$, the dropping head strategy gets higher results on the SQuAD task and comparable results on GLUE.
% Meanwhile, for $\text{WID}_{\text{11}}$, 
% For $\text{WID}_{\text{11}}^{dim}$, the performances are worse than
% These two strategies get comparable performances given same parameters.
For attention heads in $\text{WID}_{\text{55}}$, both 43 and 64 are large enough to encode the textual information in the representation subspace.
Thus, the $\text{WID}_{\text{55}}^{dim}$ with more attention heads gets slightly better results.
Similarly, the attention heads with size 16 perform worse due to the limited representation subspace, leading to the poor performance of $\text{WID}_{\text{11}}^{dim}$.

\subsection{Impact of Teacher Models}
To study the impact of teacher models, we compare the results of three teachers, including 1) $\text{BERT}_{\text{base}}$, 
2) $\text{WID}_{\text{55}}^{head}$, which is compressed by $\text{BERT}_{\text{base}}$ adopting the dropping head strategy,
3) $\text{BERT}_{\text{55}}$, which shares the same architecture as $\text{WID}_{\text{55}}^{head}$.
Both $\text{BERT}_{\text{base}}$ and $\text{BERT}_{\text{55}}$ are downloaded from the official repository \footnote{https://github.com/google-research/bert}.
We compress these three teachers to $\text{WID}_{\text{11}}^{head}$ employing the dropping head strategy.
Table \ref{tab: distill_over_gene} shows the results of three teachers.
Some findings are summarized as follows:

(1)~A smaller teacher can also teach a smart student.
Both $\text{BERT}_{\text{base}}$ and $\text{BERT}_{\text{55}}$ are pre-trained on the MLM tasks.
But the student from $\text{BERT}_{\text{55}}$ gets an average score of 77.5, which is comparable to 77.6 from the student of $\text{BERT}_{\text{base}}$.
A similar conclusion is also observed in \citet{DBLP:conf/acl/ZhangYLWXWS23}.

(2)~An educated teacher teaches better.
The $\text{WID}_{\text{55}}^{head}$ is compressed by $\text{BERT}_{\text{base}}$ adopting the dropping head strategy.
Compared to $\text{BERT}_{\text{55}}$ under the same architecture, $\text{WID}_{\text{55}}^{head}$ can teach a better student on both GLUE benchmarks and the SQuAD task.

\subsection{Looking into WID}

\begin{figure*}[t]
	\centering
	\includegraphics[width=\linewidth]{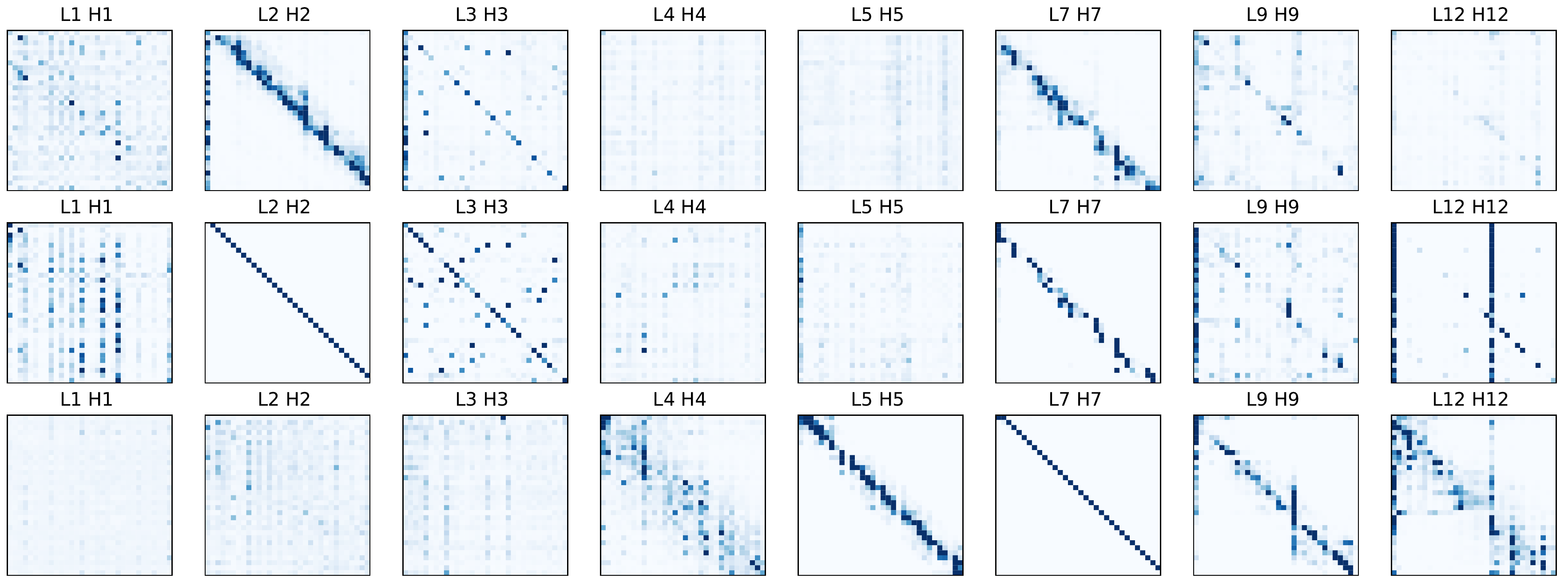}
	\caption{
	Attention distributions under same input tokens for $\text{BERT}_{\text{base}}$~(upper), $\text{WID}_{\text{11}}^{dim}$~(middle), and $\text{BERT}_{\text{11}}$~(bottom).
	Our WID can learn the knowledge about attention distributions from the teacher without any alignment loss.
	}
	\label{atten_di}
\end{figure*}

% We also investigate whether WID can learn the knowledge such as attention patterns from teacher model.
We visualize the attention distributions between the teacher $\text{BERT}_{\text{base}}$ and the student $\text{WID}_{\text{11}}^{dim}$ with the same input tokens.
For more comparison, we also pre-train $\text{BERT}_{\text{11}}$ from scratch which shares the same architecture as $\text{WID}_{\text{11}}^{dim}$.
As shown in Figure \ref{atten_di}, WID can learn the attention patterns in various layers of the teacher model $\text{BERT}_{\text{base}}$, while $\text{BERT}_{\text{11}}$ can not.
% is much more different.
The results of more attention heads can be found in Appendix \ref{apped: attn_dis}.

% In WID, we adopt the hard loss for the pre-training task during the distillation, without any alignment loss between the teacher model and the student model.
In WID, we do not use any alignment loss between the teacher and the student.
However, the compressed student model can still learn attention patterns.
This indicates that inheriting the weights can also inherit high-level semantic knowledge.

\section{Related Work}

\subsection{BERT Compression}
Transformer-based Pre-trained Language Models~(PLMs) can be compressed via Quantization \citep{DBLP:conf/iclr/StockFGGGJJ21, DBLP:conf/acl/TaoHZSJLLW22}, Matrix Decomposition \citep{DBLP:conf/coling/MaoWWZWZYTB20}, Pruning \citep{DBLP:conf/acl/XiaZC22,DBLP:conf/emnlp/LagunasCSR21}, and Knowledge Distillation \citep{DBLP:conf/emnlp/JiaoYSJCL0L20,DBLP:conf/nips/WangW0B0020}.
We refer the readers to \citet{DBLP:journals/corr/abs-2002-11985} for a comprehensive survey.
In this paper, we focus on KD for BERT compression.

\subsection{Knowledge Distillation}
KD aims to transfer the knowledge from the teacher model to the student model \citep{DBLP:journals/corr/HintonVD15,wang2023riformer,wu2023edge}.
The distillation methods can be directly divided into three main categories: offline distillation, online distillation, and self-distillation \citep{DBLP:journals/ijcv/GouYMT21}.
For PLMs, the majority of methods follow the offline distillation pattern where the teacher model is pre-trained before distillation.
Meanwhile, distillation methods for PLMs can be divided into task-agnostic, which distills the PLM in pre-training stage, and task-specific, which fine-tunes the teacher model on specific tasks and then distills.

In this work, we focus on the task-agnostic distillation.
% since the task-specific fine-tuning procedure of large PLMs is costly and time-consuming while the task-agnostic distilled models can be directly fine-tuned on downstream tasks.
Previous methods mainly focus on designing extra matching losses for the student model to mimic the teacher model.
These losses mainly include feature-based loss for features in intermediate layers and logit-based loss for output logits. 
DistilBERT \citep{DBLP:journals/corr/abs-1910-01108} adopts the output logit and embedding outputs of the teacher to train the student.
TinyBERT \citep{DBLP:conf/emnlp/JiaoYSJCL0L20} and MobileBERT \citep{DBLP:conf/acl/SunYSLYZ20} further employ the self-attention distributions and hidden states for alignment loss.
Such layer-to-layer distillation restricts the number of student layers or requires an extra mapping function.
To address this issue, MiniLM \citep{DBLP:conf/nips/WangW0B0020} proposes a new loss based on the attention matrix and values-values scaled dot-product.
WD \citep{DBLP:conf/acl/LinLWLDXZ20} employs a similar idea to inherit the knowledge in parameters.
However, WD initializes the weights of student models randomly and still requires alignment losses.

Different from existing methods, WID does not require additional alignment losses, thus avoiding laborious selection for both loss functions and loss weights.

% In fact, WID directly leverages the knowledge contained in the weights of the teacher model.
\section{Conclusion}
This work proposes a novel Weight-Inherited Distillation~(WID) method for task-agnostic BERT compression.
In WID, we factorize the compression process as weight mappings, and then design the row compactors and column compactors for row mappings and column mappings, respectively.
% In WID, we factorize the knowledge distillation process into compression weight matrices, and design the row compactors and column compactors for row mapping and column mapping.
Empirical results on various student model sizes demonstrate the effectiveness of WID.
Further analysis indicates that inheriting the weights can also inherit high-level semantic knowledge such as attention patterns.
% In future work, we would extend our WID for depth compress in PLMs.
In future work, we would consider reducing the extra memory cost by compactor layers, such as compactor sharing.
Moreover, employing WID on the large language model~(LLM) would be another interesting topic.
% other backbones such as graph neural networks 
% would be another interesting topic.

\section*{Acknowledge}
This work was supported by the Shenzhen Science and Technology under Grant JSGG20220831110203007, and by the Theme-based Research Scheme~(TRS) project T45-701/22-R, Hong Kong SAR.

% \clearpage
\section*{Limitations}
Our proposed WID inserts row/column compactors to learn the mappings from the teacher model to the student model.
Thus, WID requires additional computational time and memory.
However, WID still outperforms TinyBERT with fewer time costs.
% We reduce the training steps for WID.
As shown in Table \ref{appendix: less},
$\text{WID}_{\text{55}}^{dim}$ trained with 100k steps achieves a higher score and saves more than 50\% time costs compared to TinyBERT.
% Actually, there is no free lunch.
However, we believe that such a trade-off is valuable since a faster and better compact student would save more time on downstream tasks. 

% \section*{Ethical Considerations}
% XXX

% \clearpage
% \bibliography{ref}

\begin{thebibliography}{46}
\expandafter\ifx\csname natexlab\endcsname\relax\def\natexlab#1{#1}\fi

\bibitem[{Ba et~al.(2016)Ba, Kiros, and Hinton}]{DBLP:journals/corr/BaKH16}
Lei~Jimmy Ba, Jamie~Ryan Kiros, and Geoffrey~E. Hinton. 2016.
\newblock \href {http://arxiv.org/abs/1607.06450} {Layer normalization}.
\newblock \emph{CoRR}, abs/1607.06450.

\bibitem[{Bentivogli et~al.(2009)Bentivogli, Magnini, Dagan, Dang, and
  Giampiccolo}]{DBLP:conf/tac/BentivogliMDDG09}
Luisa Bentivogli, Bernardo Magnini, Ido Dagan, Hoa~Trang Dang, and Danilo
  Giampiccolo. 2009.
\newblock \href
  {https://tac.nist.gov/publications/2009/additional.papers/RTE5\_overview.proceedings.pdf}
  {The fifth {PASCAL} recognizing textual entailment challenge}.
\newblock In \emph{Proceedings of the Second Text Analysis Conference, {TAC}
  2009, Gaithersburg, Maryland, USA, November 16-17, 2009}. {NIST}.

\bibitem[{Cer et~al.(2017)Cer, Diab, Agirre, Lopez{-}Gazpio, and
  Specia}]{DBLP:conf/semeval/CerDALS17}
Daniel~M. Cer, Mona~T. Diab, Eneko Agirre, I{\~{n}}igo Lopez{-}Gazpio, and
  Lucia Specia. 2017.
\newblock \href {https://doi.org/10.18653/v1/S17-2001} {Semeval-2017 task 1:
  Semantic textual similarity multilingual and crosslingual focused
  evaluation}.
\newblock In \emph{Proceedings of the 11th International Workshop on Semantic
  Evaluation, SemEval@ACL 2017, Vancouver, Canada, August 3-4, 2017}, pages
  1--14. Association for Computational Linguistics.

\bibitem[{Chen et~al.(2018)Chen, Zhang, Zhang, and Zhao}]{chen2018quora}
Zihan Chen, Hongbo Zhang, Xiaoji Zhang, and Leqi Zhao. 2018.
\newblock \href {http://static.hongbozhang.me/doc/STAT_441_Report.pdf} {Quora
  question pairs}.

\bibitem[{Devlin et~al.(2019)Devlin, Chang, Lee, and
  Toutanova}]{DBLP:conf/naacl/DevlinCLT19}
Jacob Devlin, Ming{-}Wei Chang, Kenton Lee, and Kristina Toutanova. 2019.
\newblock \href {https://doi.org/10.18653/v1/n19-1423} {{BERT:} pre-training of
  deep bidirectional transformers for language understanding}.
\newblock In \emph{Proceedings of the 2019 Conference of the North American
  Chapter of the Association for Computational Linguistics: Human Language
  Technologies, {NAACL-HLT} 2019, Minneapolis, MN, USA, June 2-7, 2019, Volume
  1 (Long and Short Papers)}, pages 4171--4186. Association for Computational
  Linguistics.

\bibitem[{Ding et~al.(2021)Ding, Hao, Tan, Liu, Han, Guo, and
  Ding}]{DBLP:conf/iccv/DingHT0HGD21}
Xiaohan Ding, Tianxiang Hao, Jianchao Tan, Ji~Liu, Jungong Han, Yuchen Guo, and
  Guiguang Ding. 2021.
\newblock \href {https://doi.org/10.1109/ICCV48922.2021.00447} {Resrep:
  Lossless {CNN} pruning via decoupling remembering and forgetting}.
\newblock In \emph{2021 {IEEE/CVF} International Conference on Computer Vision,
  {ICCV} 2021, Montreal, QC, Canada, October 10-17, 2021}, pages 4490--4500.
  {IEEE}.

\bibitem[{Dolan and Brockett(2005)}]{DBLP:conf/acl-iwp/DolanB05}
William~B. Dolan and Chris Brockett. 2005.
\newblock \href {https://aclanthology.org/I05-5002/} {Automatically
  constructing a corpus of sentential paraphrases}.
\newblock In \emph{Proceedings of the Third International Workshop on
  Paraphrasing, IWP@IJCNLP 2005, Jeju Island, Korea, October 2005, 2005}. Asian
  Federation of Natural Language Processing.

\bibitem[{Ganesh et~al.(2021)Ganesh, Chen, Lou, Khan, Yang, Chen, Winslett,
  Sajjad, and Nakov}]{DBLP:journals/corr/abs-2002-11985}
Prakhar Ganesh, Yao Chen, Xin Lou, Mohammad~Ali Khan, Yin Yang, Deming Chen,
  Marianne Winslett, Hassan Sajjad, and Preslav Nakov. 2021.
\newblock \href
  {https://direct.mit.edu/tacl/article/doi/10.1162/tacl_a_00413/107387/Compressing-Large-Scale-Transformer-Based-Models-A}
  {Compressing large-scale transformer-based models: {A} case study on {BERT}}.
\newblock \emph{Transactions of the Association for Computational Linguistics},
  9:1061--1080.

\bibitem[{Gou et~al.(2021)Gou, Yu, Maybank, and
  Tao}]{DBLP:journals/ijcv/GouYMT21}
Jianping Gou, Baosheng Yu, Stephen~J. Maybank, and Dacheng Tao. 2021.
\newblock \href {https://doi.org/10.1007/s11263-021-01453-z} {Knowledge
  distillation: {A} survey}.
\newblock \emph{Int. J. Comput. Vis.}, 129(6):1789--1819.

\bibitem[{He et~al.(2016)He, Zhang, Ren, and Sun}]{DBLP:conf/cvpr/HeZRS16}
Kaiming He, Xiangyu Zhang, Shaoqing Ren, and Jian Sun. 2016.
\newblock \href {https://doi.org/10.1109/CVPR.2016.90} {Deep residual learning
  for image recognition}.
\newblock In \emph{2016 {IEEE} Conference on Computer Vision and Pattern
  Recognition, {CVPR} 2016, Las Vegas, NV, USA, June 27-30, 2016}, pages
  770--778. {IEEE} Computer Society.

\bibitem[{Hinton et~al.(2015)Hinton, Vinyals, and
  Dean}]{DBLP:journals/corr/HintonVD15}
Geoffrey~E. Hinton, Oriol Vinyals, and Jeffrey Dean. 2015.
\newblock \href {http://arxiv.org/abs/1503.02531} {Distilling the knowledge in
  a neural network}.
\newblock \emph{CoRR}, abs/1503.02531.

\bibitem[{Hou et~al.(2020)Hou, Huang, Shang, Jiang, Chen, and
  Liu}]{DBLP:conf/nips/HouHSJCL20}
Lu~Hou, Zhiqi Huang, Lifeng Shang, Xin Jiang, Xiao Chen, and Qun Liu. 2020.
\newblock \href
  {https://proceedings.neurips.cc/paper/2020/hash/6f5216f8d89b086c18298e043bfe48ed-Abstract.html}
  {Dynabert: Dynamic {BERT} with adaptive width and depth}.
\newblock In \emph{Advances in Neural Information Processing Systems 33: Annual
  Conference on Neural Information Processing Systems 2020, NeurIPS 2020,
  December 6-12, 2020, virtual}.

\bibitem[{Jiao et~al.(2020)Jiao, Yin, Shang, Jiang, Chen, Li, Wang, and
  Liu}]{DBLP:conf/emnlp/JiaoYSJCL0L20}
Xiaoqi Jiao, Yichun Yin, Lifeng Shang, Xin Jiang, Xiao Chen, Linlin Li, Fang
  Wang, and Qun Liu. 2020.
\newblock \href {https://doi.org/10.18653/v1/2020.findings-emnlp.372}
  {Tinybert: Distilling {BERT} for natural language understanding}.
\newblock In \emph{Findings of the Association for Computational Linguistics:
  {EMNLP} 2020, Online Event, 16-20 November 2020}, volume {EMNLP} 2020 of
  \emph{Findings of {ACL}}, pages 4163--4174. Association for Computational
  Linguistics.

\bibitem[{Lagunas et~al.(2021)Lagunas, Charlaix, Sanh, and
  Rush}]{DBLP:conf/emnlp/LagunasCSR21}
Fran{\c{c}}ois Lagunas, Ella Charlaix, Victor Sanh, and Alexander~M. Rush.
  2021.
\newblock \href {https://doi.org/10.18653/v1/2021.emnlp-main.829} {Block
  pruning for faster transformers}.
\newblock In \emph{Proceedings of the 2021 Conference on Empirical Methods in
  Natural Language Processing, {EMNLP} 2021, Virtual Event / Punta Cana,
  Dominican Republic, 7-11 November, 2021}, pages 10619--10629. Association for
  Computational Linguistics.

\bibitem[{Lan et~al.(2020)Lan, Chen, Goodman, Gimpel, Sharma, and
  Soricut}]{DBLP:conf/iclr/LanCGGSS20}
Zhenzhong Lan, Mingda Chen, Sebastian Goodman, Kevin Gimpel, Piyush Sharma, and
  Radu Soricut. 2020.
\newblock \href {https://openreview.net/forum?id=H1eA7AEtvS} {{ALBERT:} {A}
  lite {BERT} for self-supervised learning of language representations}.
\newblock In \emph{8th International Conference on Learning Representations,
  {ICLR} 2020, Addis Ababa, Ethiopia, April 26-30, 2020}. OpenReview.net.

\bibitem[{LeCun et~al.(1989)LeCun, Denker, and Solla}]{DBLP:conf/nips/CunDS89}
Yann LeCun, John~S. Denker, and Sara~A. Solla. 1989.
\newblock \href {http://papers.nips.cc/paper/250-optimal-brain-damage} {Optimal
  brain damage}.
\newblock In \emph{Advances in Neural Information Processing Systems 2, {[NIPS}
  Conference, Denver, Colorado, USA, November 27-30, 1989]}, pages 598--605.
  Morgan Kaufmann.

\bibitem[{Li et~al.(2020)Li, Kong, Zhang, Li, Li, Liu, and
  Ding}]{DBLP:conf/emnlp/LiKZ0LLD20}
Bingbing Li, Zhenglun Kong, Tianyun Zhang, Ji~Li, Zhengang Li, Hang Liu, and
  Caiwen Ding. 2020.
\newblock \href {https://doi.org/10.18653/v1/2020.findings-emnlp.286}
  {Efficient transformer-based large scale language representations using
  hardware-friendly block structured pruning}.
\newblock In \emph{Findings of the Association for Computational Linguistics:
  {EMNLP} 2020, Online Event, 16-20 November 2020}, volume {EMNLP} 2020 of
  \emph{Findings of {ACL}}, pages 3187--3199. Association for Computational
  Linguistics.

\bibitem[{Lin et~al.(2021)Lin, Li, Wang, Li, Du, Xiao, and
  Zhu}]{DBLP:conf/acl/LinLWLDXZ20}
Ye~Lin, Yanyang Li, Ziyang Wang, Bei Li, Quan Du, Tong Xiao, and Jingbo Zhu.
  2021.
\newblock \href {https://doi.org/10.18653/V1/2021.ACL-LONG.162} {Weight
  distillation: Transferring the knowledge in neural network parameters}.
\newblock In \emph{Proceedings of the 59th Annual Meeting of the Association
  for Computational Linguistics and the 11th International Joint Conference on
  Natural Language Processing, {ACL/IJCNLP} 2021, (Volume 1: Long Papers),
  Virtual Event, August 1-6, 2021}, pages 2076--2088. Association for
  Computational Linguistics.

\bibitem[{Liu et~al.(2019)Liu, Ott, Goyal, Du, Joshi, Chen, Levy, Lewis,
  Zettlemoyer, and Stoyanov}]{DBLP:journals/corr/abs-1907-11692}
Yinhan Liu, Myle Ott, Naman Goyal, Jingfei Du, Mandar Joshi, Danqi Chen, Omer
  Levy, Mike Lewis, Luke Zettlemoyer, and Veselin Stoyanov. 2019.
\newblock \href {http://arxiv.org/abs/1907.11692} {Roberta: {A} robustly
  optimized {BERT} pretraining approach}.
\newblock \emph{CoRR}, abs/1907.11692.

\bibitem[{Loshchilov and Hutter(2019)}]{DBLP:conf/iclr/LoshchilovH19}
Ilya Loshchilov and Frank Hutter. 2019.
\newblock \href {https://openreview.net/forum?id=Bkg6RiCqY7} {Decoupled weight
  decay regularization}.
\newblock In \emph{7th International Conference on Learning Representations,
  {ICLR} 2019, New Orleans, LA, USA, May 6-9, 2019}. OpenReview.net.

\bibitem[{Mao et~al.(2020)Mao, Wang, Wu, Zhang, Wang, Zhang, Yang, Tong, and
  Bai}]{DBLP:conf/coling/MaoWWZWZYTB20}
Yihuan Mao, Yujing Wang, Chufan Wu, Chen Zhang, Yang Wang, Quanlu Zhang, Yaming
  Yang, Yunhai Tong, and Jing Bai. 2020.
\newblock \href {https://doi.org/10.18653/v1/2020.coling-main.287} {Ladabert:
  Lightweight adaptation of {BERT} through hybrid model compression}.
\newblock In \emph{Proceedings of the 28th International Conference on
  Computational Linguistics, {COLING} 2020, Barcelona, Spain (Online), December
  8-13, 2020}, pages 3225--3234. International Committee on Computational
  Linguistics.

\bibitem[{Rajpurkar et~al.(2016)Rajpurkar, Zhang, Lopyrev, and
  Liang}]{DBLP:conf/emnlp/RajpurkarZLL16}
Pranav Rajpurkar, Jian Zhang, Konstantin Lopyrev, and Percy Liang. 2016.
\newblock \href {https://doi.org/10.18653/v1/d16-1264} {Squad: 100, 000+
  questions for machine comprehension of text}.
\newblock In \emph{Proceedings of the 2016 Conference on Empirical Methods in
  Natural Language Processing, {EMNLP} 2016, Austin, Texas, USA, November 1-4,
  2016}, pages 2383--2392. The Association for Computational Linguistics.

\bibitem[{Reid et~al.(2021)Reid, Marrese{-}Taylor, and
  Matsuo}]{DBLP:conf/emnlp/ReidMM21}
Machel Reid, Edison Marrese{-}Taylor, and Yutaka Matsuo. 2021.
\newblock \href {https://doi.org/10.18653/v1/2021.findings-emnlp.344}
  {Subformer: Exploring weight sharing for parameter efficiency in generative
  transformers}.
\newblock In \emph{Findings of the Association for Computational Linguistics:
  {EMNLP} 2021, Virtual Event / Punta Cana, Dominican Republic, 16-20 November,
  2021}, pages 4081--4090. Association for Computational Linguistics.

\bibitem[{Sanh et~al.(2019)Sanh, Debut, Chaumond, and
  Wolf}]{DBLP:journals/corr/abs-1910-01108}
Victor Sanh, Lysandre Debut, Julien Chaumond, and Thomas Wolf. 2019.
\newblock \href {http://arxiv.org/abs/1910.01108} {Distilbert, a distilled
  version of {BERT:} smaller, faster, cheaper and lighter}.
\newblock \emph{CoRR}, abs/1910.01108.

\bibitem[{Socher et~al.(2013)Socher, Perelygin, Wu, Chuang, Manning, Ng, and
  Potts}]{DBLP:conf/emnlp/SocherPWCMNP13}
Richard Socher, Alex Perelygin, Jean Wu, Jason Chuang, Christopher~D. Manning,
  Andrew~Y. Ng, and Christopher Potts. 2013.
\newblock \href {https://aclanthology.org/D13-1170/} {Recursive deep models for
  semantic compositionality over a sentiment treebank}.
\newblock In \emph{Proceedings of the 2013 Conference on Empirical Methods in
  Natural Language Processing, {EMNLP} 2013, 18-21 October 2013, Grand Hyatt
  Seattle, Seattle, Washington, USA, {A} meeting of SIGDAT, a Special Interest
  Group of the {ACL}}, pages 1631--1642. {ACL}.

\bibitem[{Stock et~al.(2021)Stock, Fan, Graham, Grave, Gribonval, J{\'{e}}gou,
  and Joulin}]{DBLP:conf/iclr/StockFGGGJJ21}
Pierre Stock, Angela Fan, Benjamin Graham, Edouard Grave, R{\'{e}}mi Gribonval,
  Herv{\'{e}} J{\'{e}}gou, and Armand Joulin. 2021.
\newblock \href {https://openreview.net/forum?id=dV19Yyi1fS3} {Training with
  quantization noise for extreme model compression}.
\newblock In \emph{9th International Conference on Learning Representations,
  {ICLR} 2021, Virtual Event, Austria, May 3-7, 2021}. OpenReview.net.

\bibitem[{Sun et~al.(2019)Sun, Cheng, Gan, and Liu}]{DBLP:conf/emnlp/SunCGL19}
Siqi Sun, Yu~Cheng, Zhe Gan, and Jingjing Liu. 2019.
\newblock \href {https://doi.org/10.18653/v1/D19-1441} {Patient knowledge
  distillation for {BERT} model compression}.
\newblock In \emph{Proceedings of the 2019 Conference on Empirical Methods in
  Natural Language Processing and the 9th International Joint Conference on
  Natural Language Processing, {EMNLP-IJCNLP} 2019, Hong Kong, China, November
  3-7, 2019}, pages 4322--4331. Association for Computational Linguistics.

\bibitem[{Sun et~al.(2020)Sun, Yu, Song, Liu, Yang, and
  Zhou}]{DBLP:conf/acl/SunYSLYZ20}
Zhiqing Sun, Hongkun Yu, Xiaodan Song, Renjie Liu, Yiming Yang, and Denny Zhou.
  2020.
\newblock \href {https://doi.org/10.18653/v1/2020.acl-main.195} {Mobilebert: a
  compact task-agnostic {BERT} for resource-limited devices}.
\newblock In \emph{Proceedings of the 58th Annual Meeting of the Association
  for Computational Linguistics, {ACL} 2020, Online, July 5-10, 2020}, pages
  2158--2170. Association for Computational Linguistics.

\bibitem[{Tao et~al.(2022)Tao, Hou, Zhang, Shang, Jiang, Liu, Luo, and
  Wong}]{DBLP:conf/acl/TaoHZSJLLW22}
Chaofan Tao, Lu~Hou, Wei Zhang, Lifeng Shang, Xin Jiang, Qun Liu, Ping Luo, and
  Ngai Wong. 2022.
\newblock \href {https://aclanthology.org/2022.acl-long.331} {Compression of
  generative pre-trained language models via quantization}.
\newblock In \emph{Proceedings of the 60th Annual Meeting of the Association
  for Computational Linguistics (Volume 1: Long Papers), {ACL} 2022, Dublin,
  Ireland, May 22-27, 2022}, pages 4821--4836. Association for Computational
  Linguistics.

\bibitem[{Vaswani et~al.(2017)Vaswani, Shazeer, Parmar, Uszkoreit, Jones,
  Gomez, Kaiser, and Polosukhin}]{DBLP:conf/nips/VaswaniSPUJGKP17}
Ashish Vaswani, Noam Shazeer, Niki Parmar, Jakob Uszkoreit, Llion Jones,
  Aidan~N. Gomez, Lukasz Kaiser, and Illia Polosukhin. 2017.
\newblock \href
  {https://proceedings.neurips.cc/paper/2017/hash/3f5ee243547dee91fbd053c1c4a845aa-Abstract.html}
  {Attention is all you need}.
\newblock In \emph{Advances in Neural Information Processing Systems 30: Annual
  Conference on Neural Information Processing Systems 2017, December 4-9, 2017,
  Long Beach, CA, {USA}}, pages 5998--6008.

\bibitem[{Wang et~al.(2019)Wang, Singh, Michael, Hill, Levy, and
  Bowman}]{DBLP:conf/iclr/WangSMHLB19}
Alex Wang, Amanpreet Singh, Julian Michael, Felix Hill, Omer Levy, and
  Samuel~R. Bowman. 2019.
\newblock \href {https://openreview.net/forum?id=rJ4km2R5t7} {{GLUE:} {A}
  multi-task benchmark and analysis platform for natural language
  understanding}.
\newblock In \emph{7th International Conference on Learning Representations,
  {ICLR} 2019, New Orleans, LA, USA, May 6-9, 2019}. OpenReview.net.

\bibitem[{Wang et~al.(2023)Wang, Zhang, Liu, Wu, Yang, Liu, Chen, Luo, and
  Lin}]{wang2023riformer}
Jiahao Wang, Songyang Zhang, Yong Liu, Taiqiang Wu, Yujiu Yang, Xihui Liu, Kai
  Chen, Ping Luo, and Dahua Lin. 2023.
\newblock Riformer: Keep your vision backbone effective but removing token
  mixer.
\newblock In \emph{Proceedings of the IEEE/CVF Conference on Computer Vision
  and Pattern Recognition}, pages 14443--14452.

\bibitem[{Wang et~al.(2021)Wang, Bao, Huang, Dong, and
  Wei}]{DBLP:conf/acl/WangBHDW21}
Wenhui Wang, Hangbo Bao, Shaohan Huang, Li~Dong, and Furu Wei. 2021.
\newblock \href {https://doi.org/10.18653/v1/2021.findings-acl.188} {Minilmv2:
  Multi-head self-attention relation distillation for compressing pretrained
  transformers}.
\newblock In \emph{Findings of the Association for Computational Linguistics:
  {ACL/IJCNLP} 2021, Online Event, August 1-6, 2021}, volume {ACL/IJCNLP} 2021
  of \emph{Findings of {ACL}}, pages 2140--2151. Association for Computational
  Linguistics.

\bibitem[{Wang et~al.(2020)Wang, Wei, Dong, Bao, Yang, and
  Zhou}]{DBLP:conf/nips/WangW0B0020}
Wenhui Wang, Furu Wei, Li~Dong, Hangbo Bao, Nan Yang, and Ming Zhou. 2020.
\newblock \href
  {https://proceedings.neurips.cc/paper/2020/hash/3f5ee243547dee91fbd053c1c4a845aa-Abstract.html}
  {Minilm: Deep self-attention distillation for task-agnostic compression of
  pre-trained transformers}.
\newblock In \emph{Advances in Neural Information Processing Systems 33: Annual
  Conference on Neural Information Processing Systems 2020, NeurIPS 2020,
  December 6-12, 2020, virtual}.

\bibitem[{Warstadt et~al.(2019)Warstadt, Singh, and
  Bowman}]{DBLP:journals/tacl/WarstadtSB19}
Alex Warstadt, Amanpreet Singh, and Samuel~R. Bowman. 2019.
\newblock \href {https://doi.org/10.1162/tacl\_a\_00290} {Neural network
  acceptability judgments}.
\newblock \emph{Trans. Assoc. Comput. Linguistics}, 7:625--641.

\bibitem[{Williams et~al.(2018)Williams, Nangia, and
  Bowman}]{DBLP:conf/naacl/WilliamsNB18}
Adina Williams, Nikita Nangia, and Samuel~R. Bowman. 2018.
\newblock \href {https://doi.org/10.18653/v1/n18-1101} {A broad-coverage
  challenge corpus for sentence understanding through inference}.
\newblock In \emph{Proceedings of the 2018 Conference of the North American
  Chapter of the Association for Computational Linguistics: Human Language
  Technologies, {NAACL-HLT} 2018, New Orleans, Louisiana, USA, June 1-6, 2018,
  Volume 1 (Long Papers)}, pages 1112--1122. Association for Computational
  Linguistics.

\bibitem[{Wu et~al.(2023)Wu, Zhao, Wang, Bai, Wang, Wong, and
  Yang}]{wu2023edge}
Taiqiang Wu, Zhe Zhao, Jiahao Wang, Xingyu Bai, Lei Wang, Ngai Wong, and Yujiu
  Yang. 2023.
\newblock Edge-free but structure-aware: Prototype-guided knowledge
  distillation from gnns to mlps.
\newblock \emph{arXiv preprint arXiv:2303.13763}.

\bibitem[{Wu et~al.(2016)Wu, Schuster, Chen, Le, Norouzi, Macherey, Krikun,
  Cao, Gao, Macherey, Klingner, Shah, Johnson, Liu, Kaiser, Gouws, Kato, Kudo,
  Kazawa, Stevens, Kurian, Patil, Wang, Young, Smith, Riesa, Rudnick, Vinyals,
  Corrado, Hughes, and Dean}]{DBLP:journals/corr/WuSCLNMKCGMKSJL16}
Yonghui Wu, Mike Schuster, Zhifeng Chen, Quoc~V. Le, Mohammad Norouzi, Wolfgang
  Macherey, Maxim Krikun, Yuan Cao, Qin Gao, Klaus Macherey, Jeff Klingner,
  Apurva Shah, Melvin Johnson, Xiaobing Liu, Lukasz Kaiser, Stephan Gouws,
  Yoshikiyo Kato, Taku Kudo, Hideto Kazawa, Keith Stevens, George Kurian,
  Nishant Patil, Wei Wang, Cliff Young, Jason Smith, Jason Riesa, Alex Rudnick,
  Oriol Vinyals, Greg Corrado, Macduff Hughes, and Jeffrey Dean. 2016.
\newblock \href {http://arxiv.org/abs/1609.08144} {Google's neural machine
  translation system: Bridging the gap between human and machine translation}.
\newblock \emph{CoRR}, abs/1609.08144.

\bibitem[{Xia et~al.(2022)Xia, Zhong, and Chen}]{DBLP:conf/acl/XiaZC22}
Mengzhou Xia, Zexuan Zhong, and Danqi Chen. 2022.
\newblock \href {https://aclanthology.org/2022.acl-long.107} {Structured
  pruning learns compact and accurate models}.
\newblock In \emph{Proceedings of the 60th Annual Meeting of the Association
  for Computational Linguistics (Volume 1: Long Papers), {ACL} 2022, Dublin,
  Ireland, May 22-27, 2022}, pages 1513--1528. Association for Computational
  Linguistics.

\bibitem[{Xu et~al.(2020)Xu, Zhou, Ge, Wei, and
  Zhou}]{DBLP:conf/emnlp/XuZGWZ20}
Canwen Xu, Wangchunshu Zhou, Tao Ge, Furu Wei, and Ming Zhou. 2020.
\newblock \href {https://doi.org/10.18653/v1/2020.emnlp-main.633}
  {Bert-of-theseus: Compressing {BERT} by progressive module replacing}.
\newblock In \emph{Proceedings of the 2020 Conference on Empirical Methods in
  Natural Language Processing, {EMNLP} 2020, Online, November 16-20, 2020},
  pages 7859--7869. Association for Computational Linguistics.

\bibitem[{Yang et~al.(2019)Yang, Dai, Yang, Carbonell, Salakhutdinov, and
  Le}]{DBLP:conf/nips/YangDYCSL19}
Zhilin Yang, Zihang Dai, Yiming Yang, Jaime~G. Carbonell, Ruslan Salakhutdinov,
  and Quoc~V. Le. 2019.
\newblock \href
  {https://proceedings.neurips.cc/paper/2019/hash/dc6a7e655d7e5840e66733e9ee67cc69-Abstract.html}
  {Xlnet: Generalized autoregressive pretraining for language understanding}.
\newblock In \emph{Advances in Neural Information Processing Systems 32: Annual
  Conference on Neural Information Processing Systems 2019, NeurIPS 2019,
  December 8-14, 2019, Vancouver, BC, Canada}, pages 5754--5764.

\bibitem[{Zhang et~al.(2023)Zhang, Yang, Liu, Wang, Xian, Wang, and
  Song}]{DBLP:conf/acl/ZhangYLWXWS23}
Chen Zhang, Yang Yang, Jiahao Liu, Jingang Wang, Yunsen Xian, Benyou Wang, and
  Dawei Song. 2023.
\newblock \href {https://doi.org/10.18653/V1/2023.ACL-LONG.249} {Lifting the
  curse of capacity gap in distilling language models}.
\newblock In \emph{Proceedings of the 61st Annual Meeting of the Association
  for Computational Linguistics (Volume 1: Long Papers), {ACL} 2023, Toronto,
  Canada, July 9-14, 2023}, pages 4535--4553. Association for Computational
  Linguistics.

\bibitem[{Zhang et~al.(2019)Zhang, Song, Gao, Chen, Bao, and
  Ma}]{DBLP:conf/iccv/ZhangSGCBM19}
Linfeng Zhang, Jiebo Song, Anni Gao, Jingwei Chen, Chenglong Bao, and Kaisheng
  Ma. 2019.
\newblock \href {https://doi.org/10.1109/ICCV.2019.00381} {Be your own teacher:
  Improve the performance of convolutional neural networks via self
  distillation}.
\newblock In \emph{2019 {IEEE/CVF} International Conference on Computer Vision,
  {ICCV} 2019, Seoul, Korea (South), October 27 - November 2, 2019}, pages
  3712--3721. {IEEE}.

\bibitem[{Zhao et~al.(2023)Zhao, Li, Hou, Zhao, Tian, Liu, Chen, Sun, Liu, Mao
  et~al.}]{zhao2023tencentpretrain}
Zhe Zhao, Yudong Li, Cheng Hou, Jing Zhao, Rong Tian, Weijie Liu, Yiren Chen,
  Ningyuan Sun, Haoyan Liu, Weiquan Mao, et~al. 2023.
\newblock Tencentpretrain: A scalable and flexible toolkit for pre-training
  models of different modalities.
\newblock In \emph{Proceedings of the 61st Annual Meeting of the Association
  for Computational Linguistics (Volume 3: System Demonstrations)}, pages
  217--225.

\bibitem[{Zhou et~al.(2022)Zhou, Xu, and McAuley}]{DBLP:conf/acl/ZhouXM22}
Wangchunshu Zhou, Canwen Xu, and Julian~J. McAuley. 2022.
\newblock \href {https://aclanthology.org/2022.acl-long.485} {{BERT} learns to
  teach: Knowledge distillation with meta learning}.
\newblock In \emph{Proceedings of the 60th Annual Meeting of the Association
  for Computational Linguistics (Volume 1: Long Papers), {ACL} 2022, Dublin,
  Ireland, May 22-27, 2022}, pages 7037--7049. Association for Computational
  Linguistics.

\bibitem[{Zhu et~al.(2015)Zhu, Kiros, Zemel, Salakhutdinov, Urtasun, Torralba,
  and Fidler}]{DBLP:conf/iccv/ZhuKZSUTF15}
Yukun Zhu, Ryan Kiros, Richard~S. Zemel, Ruslan Salakhutdinov, Raquel Urtasun,
  Antonio Torralba, and Sanja Fidler. 2015.
\newblock \href {https://doi.org/10.1109/ICCV.2015.11} {Aligning books and
  movies: Towards story-like visual explanations by watching movies and reading
  books}.
\newblock In \emph{2015 {IEEE} International Conference on Computer Vision,
  {ICCV} 2015, Santiago, Chile, December 7-13, 2015}, pages 19--27. {IEEE}
  Computer Society.

\end{thebibliography}
% \bibliographystyle{acl_natbib}

% \clearpage
\clearpage

\appendix

\label{sec:appendix}

\section{GLUE and SQuAD}
\label{app:data}

\subsection{Data Statistics}

Table \ref{datasets} shows the sizes of the train/development set and the metrics for downstream tasks.

\begin{table}[h]
\centering
\resizebox{0.85\columnwidth}{!}{%
\begin{tabular}{lrrr} 
\toprule
{\bf Task}  & {\bf \#Train} & {\bf \#Dev} & {\bf Metric} \\ 
\midrule
SST-2 & 67k  & 872 & Accuracy      \\
QNLI  & 105k & 5.5k & Accuracy       \\
MNLI  & 393k & 20k & Accuracy      \\
QQP   & 364k & 40k & Accuracy \\
CoLA  & 8.5k & 1k & Matthews corr.\\
RTE   & 2.5k & 276 & Accuracy         \\
STS-B & 7k &   1.5k & Spearman corr. \\
MRPC  & 3.7k  & 408 & Accuracy        \\
SQuAD & 87.6k   & 34.7k & F1 \& EM     \\ 
%SQuAD & $87599$   & 34726 &F1 \& EM     \\ 
\bottomrule
\end{tabular}
}
\caption{Data statistics of GLUE and SQuAD datasets.}
\label{datasets}
\end{table}

\subsection{Hyperparameters}
We employ the grid search to fine-tune the GLUE benchmarks and SQuAD.
\paragraph{GLUE} 
The learning rate is searched in \{1e-5, 2e-5, 3e-5\}.
We set the search space for the training batch size based on the size of the training set. 
For large datasets including QNLI, MNLI, and QQP, the batch size is searched in \{32, 48\}.
For small datasets including MRPC, RTE, CoLA, and STS-B, the batch size is searched in \{4, 6\}.
For SST-2, the batch size is searched in \{8, 16\}.
All tasks are trained for 10 epochs.

\paragraph{SQuAD}
The learning rate is searched in \{1e-5, 2e-5, 3e-5\} and batch size is searched in \{4,6,8\}.
The training epochs are set to 3.

\section{Method Details}
\subsection{Algorithm}

More details about the proposed WID can be found in Algorithm \ref{alg1}.

\begin{algorithm}[!h]
    \small
% 	\vskip -0.03in
	\caption{Weight-Inherited Distillation}
	\label{alg1}
	
	{\bf Input:} teacher model $\mathcal{T}$ with width $d_t$
	
	{\bf Params:} $k$: number of rows/columns to compress,
	$N$: steps to increase $k$, $d$: increment for $k$ each time

	{\bf Output:} student model $\mathcal{S}$ with width $d_s$
	
	\begin{algorithmic}[1]
	    \STATE Add compactors for $\mathcal{T}$ to construct the re-parameterized teacher model $\mathcal{\hat{T}}$. Initialize the weights for compactors as identity matrices.
	    \STATE $k \gets 0$ ; ${M} \gets \left[ \ \right]$
		\FOR{$i=0$ to max training steps}
		\STATE Forward a batch through $\mathcal{\hat{T}}$, derive the gradients $g_{ori}$ for compactors to update
		\IF{ $i \% N == 0 \ \& \ k<d_t-d_s$}
		\STATE Calculate p-norm values
		\STATE Select the top-$k$ row/column with the lower norm to get $M$
		\STATE Get penalty gradients $g_{pen}$ following Eq. \ref{eq_pe}
		\STATE $g_{fused} \gets f(g_{ori}, g_{pen}, M)$ following Eq. \ref{eq_fuse}
		\STATE $k \gets k+ d$
		\ENDIF
		\STATE Update the compactors with corresponding $g_{fused}$ and original layers with $g_{ori}$
		\STATE Apply the compactor aligning strategy
		\ENDFOR
		\STATE Compress the compactors following Eq. \ref{eq_prune}
		\STATE Merge the compactors and  original layers following Eq. \ref{eq_merge} to get compact layers for $\mathcal{S}$
		\STATE {\bf return} $\mathcal{S}$
	\end{algorithmic}
\end{algorithm}

\subsection{Groups of Aligned Compactors}
\label{appendix-align}
Specifically, we can divide all the compactors in BERT into the following aligned groups:
\begin{itemize}
    \item One group in blue: \{CC for embedding layer, blue compactors in each Transformer layer, RC for output layer\},
    \item $L$ groups in orange: \{orange compactors in layer 1\}; \{orange compactors in layer 2\}; ... \{orange compactors in layer $L$\},
    \item $L$ groups in green:  \{green compactors in layer 1\}; \{green compactors in layer 2\}; ... \{green compactors in layer $L$\},
\end{itemize}
Where RC/CC denotes the row/column compactor and \{$\cdot$\} denotes a group.
For the only one group in blue, we calculate the column compactor for the embedding layer and duplicate~(or, flip) it for the other compactors.
For each group in orange, we calculate the column compactor for the Value projection and duplicate~(or, flip) it for the rest three compactors.
For each group in green, we calculate the column compactor for the Up-project and flip it for the other one.

\section{Extensive Analysis}

\subsection{Comparison with Task-Specific Distillation}
\label{appen: task-spec}
% It can be unfair to directly compare task-agnostic WID with task-specific distillation methods since the teacher model in task-specific distillation methods is fine-tuned for the task before distillation.
We also compare WID with task-specific distillation methods where the teacher model in task-specific distillation methods is fine-tuned for the task before distillation.
% We compare our WID with 
% 
For baselines, we select BERT-of-Theseus \citep{DBLP:conf/emnlp/XuZGWZ20}, 
DynaBERT \citep{DBLP:conf/nips/HouHSJCL20} and MetaDistill\citep{DBLP:conf/acl/ZhouXM22}.
% Moreover, we also perform the task-specific distillation for TinyBERT.
As shown in Table \ref{appendix: task}, WID also outperforms these task-specific methods on the GLUE benchmarks.

\begin{table*}[!t]
\centering
% \tableindent 
\resizebox{2.1\columnwidth}{!}
{%
    \begin{tabular}{l|cc|cccccccc|c}
        \toprule
        {\bf Method} & {\bf Type} & {\bf Params} & {\bf SST-2} & {\bf CoLA} & {\bf MRPC} & {\bf QNLI} & {\bf QQP} & {\bf RTE} & {\bf STS-B} & {\bf MNLI} & {\bf AVG} \\
        \midrule
        $\text{BERT}_{\text{base}}$ & Teacher & 110.1M & 92.7 & 59.1 & 90.4 & 91.7 & 91.4 & 70.8 & 90.1 & 84.5 &  83.8 \\
        \midrule
        % BERT-of-Theseus & 11.9B & 67.5M & 91.5 & 51.1 & 89.0 & 89.5 & 89.6 & 68.2 & 88.7 & 82.3 & 81.2 \\
        DynaBERT & TS-KD & 67.5M & 92.7 & 54.6 & 85.0 & 90.6 & 91.1 & 66.1 & 88.6 & 83.7 & 81.6 \\
        MetaDistill & TS-KD & 67.5M & 92.3 & 58.6 & 86.8 & 90.4 & 91.0 & 69.4 & 89.1 & 83.8 & 82.7 \\
        TinyBERT$^{*}$ & TS-KD & 67.5M & 91.9 & 52.4 & 86.5 & 89.8 & 90.6 & 67.7 & 88.7 & 83.8 & 81.4 \\
        BlockPruning & Pruning & 77.0M & 89.3 & 52.6 & 88.3 & 88.2 & 90.7 & 63.9 & 84.6 & 82.9 & 80.1 \\
        \rowcolor{gray!15} $\text{WID}_{\text{55}}$~(ours)    & TA-KD & 54.9M &  92.4 & 61.7 & 88.2 & 90.1 & 91.0 & 70.4 & 87.9 & 82.9 & 83.4 \\
        \midrule
        CoFi & Pruning & 28.4M & 90.6 & 35.6 & 82.6 & 86.1 & 90.1 & 64.7 & 83.1 & 80.6 & 76.6 \\
        % \midrule
        \rowcolor{gray!15} $\text{WID}_{\text{11}}$~(ours) & TA-KD & 11.3M &  88.8 & 44.2 & 81.9 & 85.4 & 89.5 & 60.3 & 84.5 & 78.4 & 76.6 \\
        \bottomrule
    \end{tabular}
}
\caption{
Comparison among WID, task-specific distillation methods, and pruning methods on GLUE benchmarks without data augmentation.
TS-KD and TA-KD denote task-specific knowledge distillation and task-agnostic knowledge distillation, respectively.
$^*$ means the results are taken from \citet{DBLP:conf/acl/ZhouXM22}. 
Other results are taken from the corresponding papers.
}
\label{appendix: task}
\end{table*}

\subsection{Comparison with Pruning}
\label{appendix: pruning}
We try to compare WID with pruning methods for BERT compression, including task-specific CoFi~(Coarse- and Fine-grained Pruning,\cite{DBLP:conf/acl/XiaZC22}) and BlockPruning\cite{DBLP:conf/emnlp/LiKZ0LLD20}.
% is a task-specific structured pruning method. 
% The results are taken from the original paper \cite{DBLP:conf/acl/XiaZC22}.
As mentioned in Appendix \ref{appen: task-spec}, the task-agnostic setting is more difficult than the task-specific setting.
However, as shown in Table \ref{appendix: task}, WID still achieves comparable results with less than 50\% parameters compared to CoFi, and achieves better performance than BlockPruning with 28.7\% fewer parameters.

\subsection{Less Training Steps}

In Table \ref{tab: main_res}, we report the results of $\text{WID}_{\text{55}}^{dim}$ trained for 400k steps.
We re-implement TinyBERT and train 3 epochs following the setting in \citet{DBLP:conf/emnlp/JiaoYSJCL0L20}.
We reduce the training steps for $\text{WID}_{\text{55}}^{dim}$ to 50k and 100k.
All experiments are carried out with 8 A100 GPUs.
As shown in Table \ref{appendix: less}, $\text{WID}_{\text{55}}^{dim}$ trained with 100k steps can still outperform TinyBERT and save more than 50\% training time.

\begin{table}[h]
\centering
\resizebox{0.8\columnwidth}{!}{%
\begin{tabular}{lrrr} 
\toprule
{\bf Model} & {\bf Steps} & {\bf Time} & {\bf Score} \\
\midrule
TinyBERT~(GD) & 450k & 33h & 81.27 \\
$\text{WID}_{\text{55}}^{dim}$ &50k & 8h & 80.78 \\
$\text{WID}_{\text{55}}^{dim}$ & 100k & 16h & 81.65 \\
\rowcolor{gray!15} $\text{WID}_{\text{55}}^{dim}$ & 400k & 64h & 83.08 \\
\bottomrule
\end{tabular}
}
\caption{
Comparison between TinyBERT and WID trained with less steps on GLUE benchmarks.
}
\label{appendix: less}
\end{table}

\subsection{WID for GPT Compression}
\label{appendix: gpt_compress}
\begin{figure}[h]
	\centering
	\includegraphics[width=\linewidth]{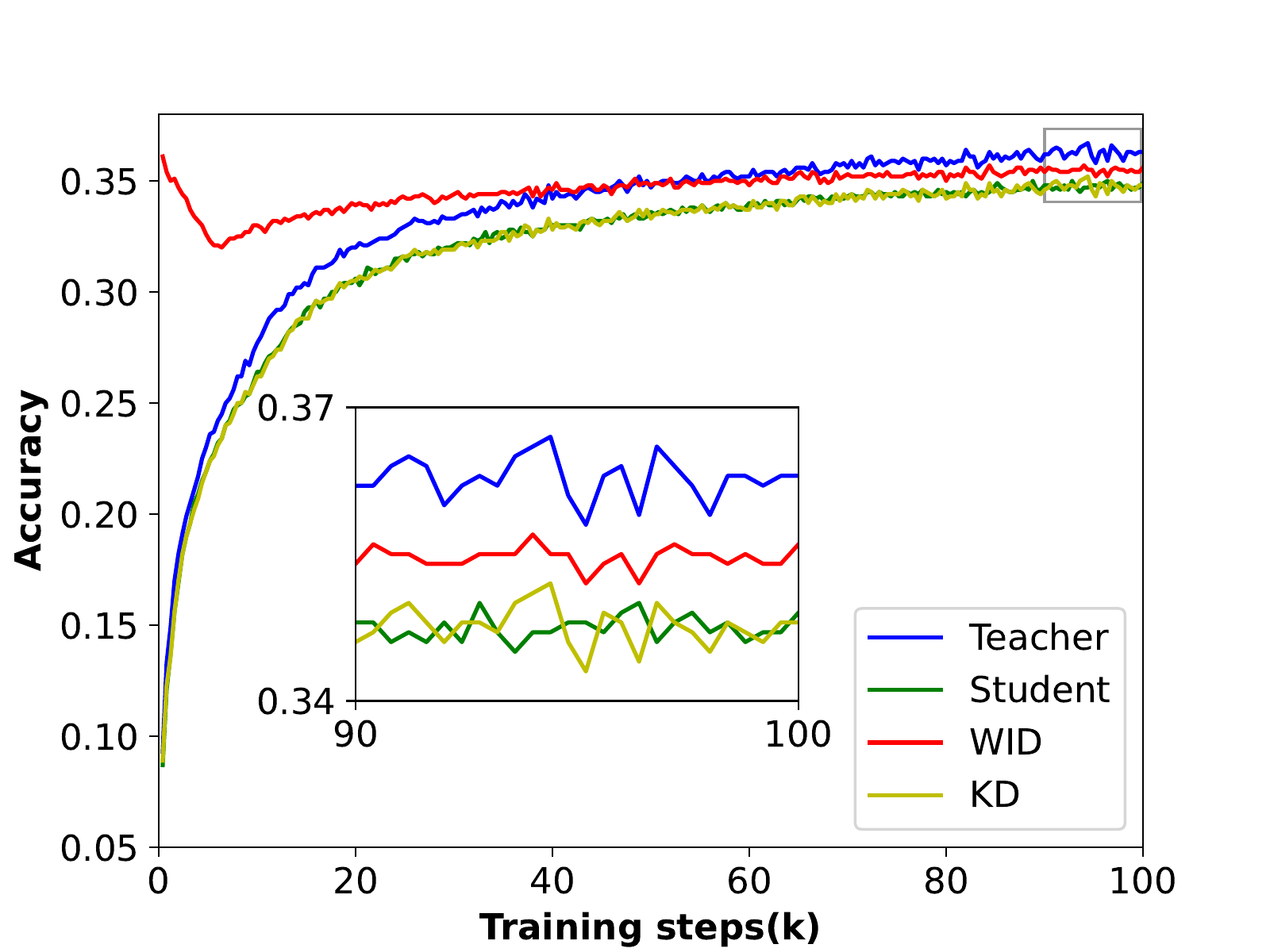}
	\caption{
	The training process for teacher GPT, vanilla student GPT, and students via KD and WID.
	}
	\label{appendx: gpt_compress}
\end{figure}

To evaluate the performance of WID on the generative pre-trained language model, we train a GPT model and compress it via vanilla KD and WID.
Due to the limited GPU memory, we train a GPT teacher~(12 layers and hidden size as 768) for 100k steps.
After that, we train a student model~(12 layers and hidden size as 512) and compress the teacher model into such a setting via vanilla KD and WID.
During distillation, we employ BookCorpus as training datasets and report the training accuracy. 
For hyperparameters, the batch size is 64 and the learning rate is 1e-4.
Figure \ref{appendx: gpt_compress} shows the training process.
We can conclude that WID still works for generative pre-trained language models, and can get better performance than vanilla KD. 
% Moreover, we leave it for future work to apply WID for larger GPT models and longer training steps.

\subsection{Attention Distributions}
\label{apped: attn_dis}
\begin{figure*}[h]
	\centering
	\includegraphics[width=1.0\linewidth]{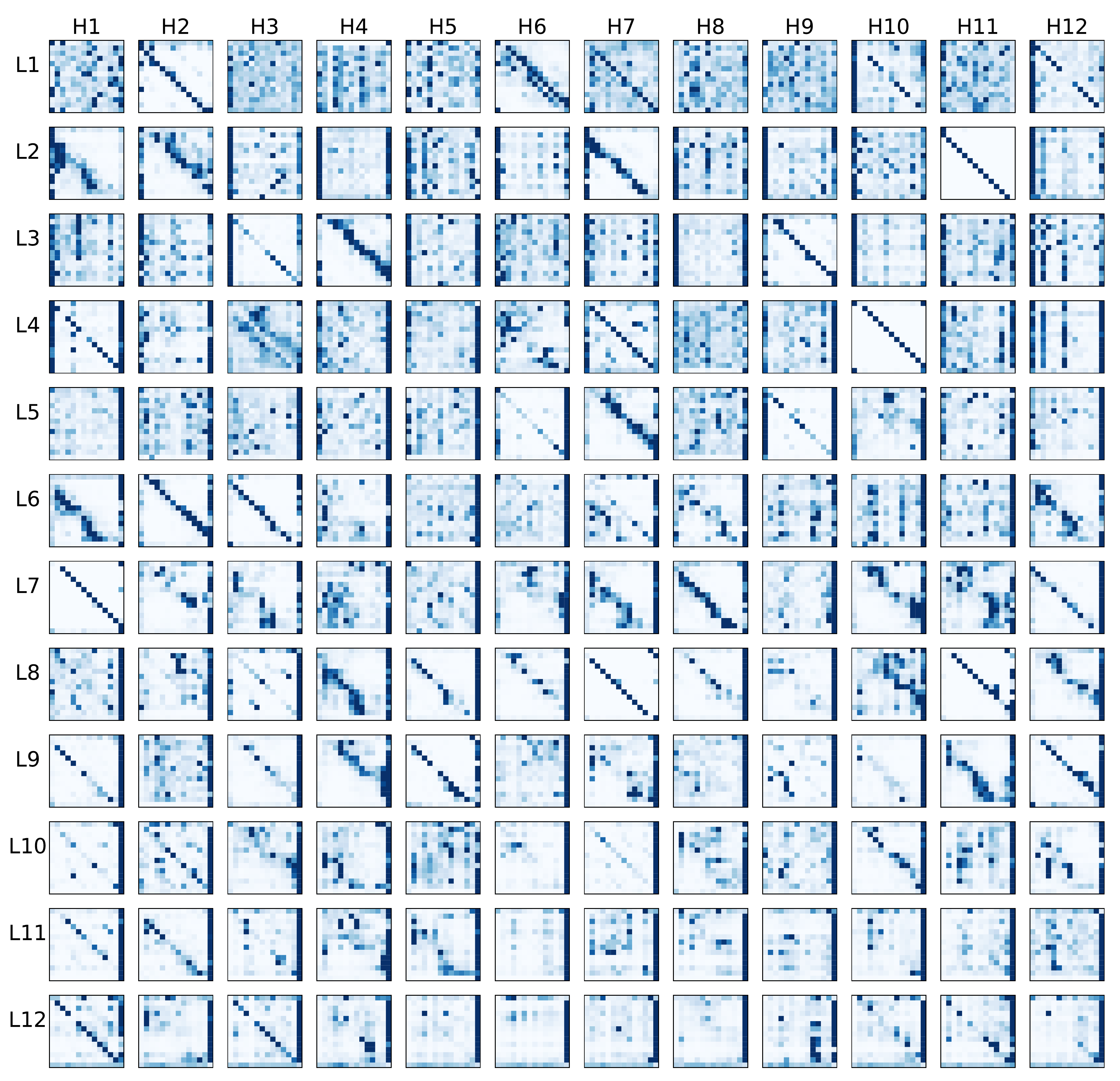}
	\caption{
	The self-attention distributions for teacher model $\text{BERT}_{\text{base}}$.
	}
	\label{appendx: teacher_attn}
\end{figure*}

\begin{figure*}[h]
	\centering
	\includegraphics[width=1.0\linewidth]{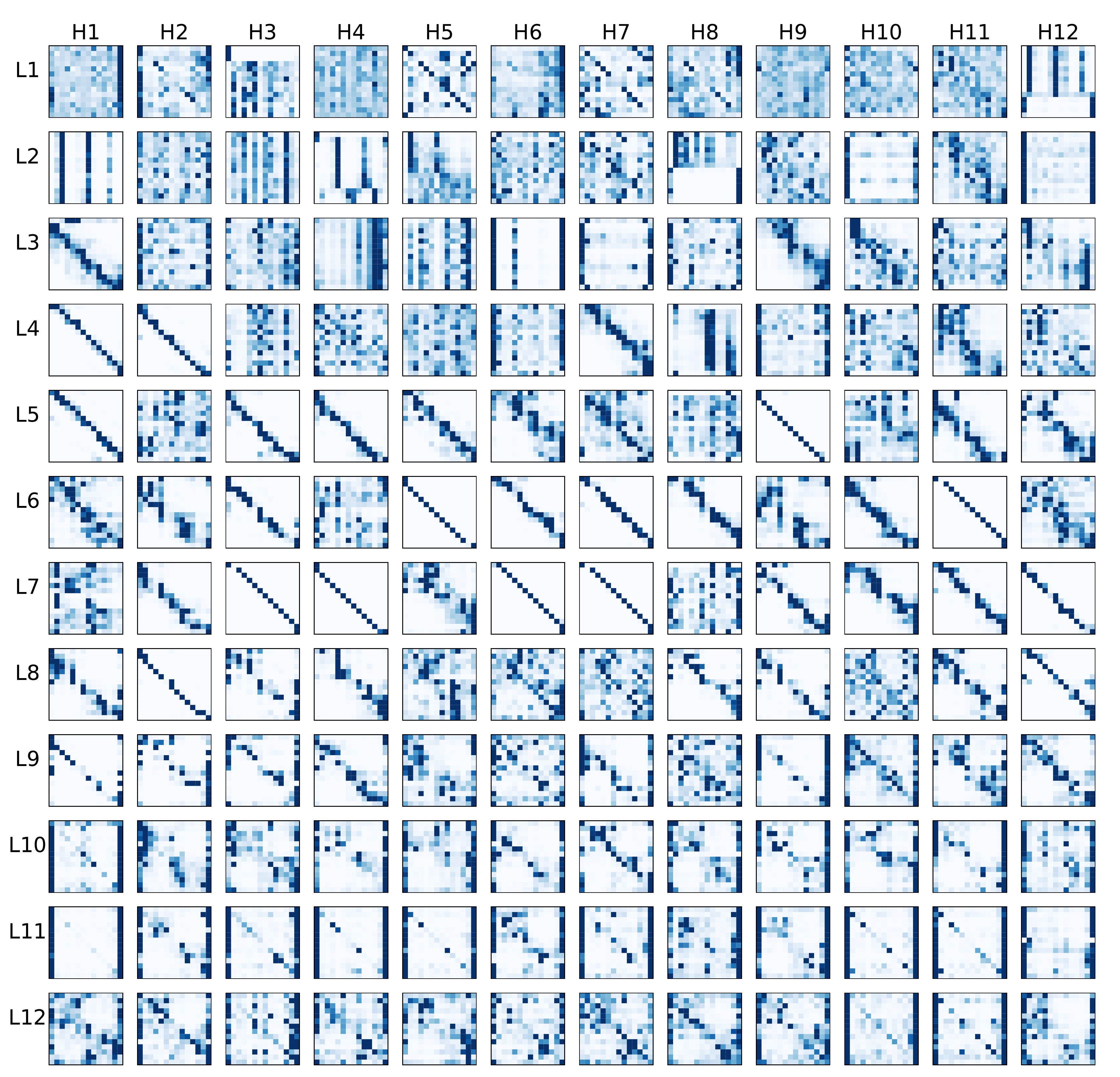}
	\caption{
	The self-attention distributions for  $\text{BERT}_{\text{11}}$.
	}
	\label{appendx: random_attn}
\end{figure*}

\begin{figure*}[h]
	\centering
	\includegraphics[width=1.0\linewidth]{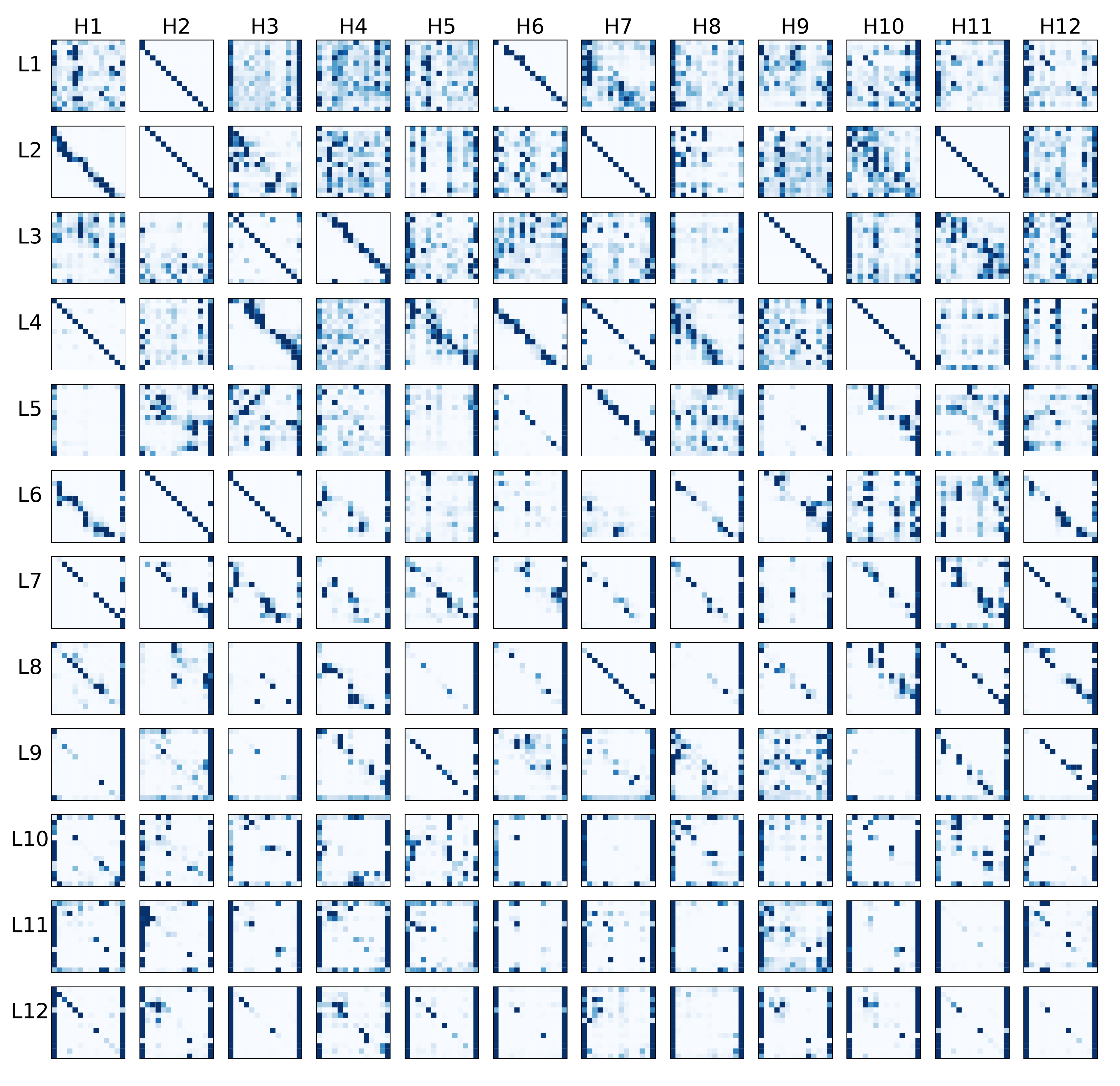}
	\caption{
	The self-attention distributions for our proposed $\text{WID}_{\text{11}}^{dim}$.
	}
	\label{appendx: student_attn}
\end{figure*}

We visualize the attention distributions for the teacher $\text{BERT}_{\text{base}}$, pre-trained $\text{BERT}_{\text{55}}$ and the student $\text{WID}_{\text{11}}^{head}$ under the same input tokens~(input sentence: "if the world harassed me, it will harass you too.") in Figure \ref{appendx: teacher_attn}, Figure \ref{appendx: random_attn} and Figure \ref{appendx: student_attn}, respectively.
% From the bottom layer to the top layer, 
WID can effectively learn the attention patterns from the teacher model while $\text{BERT}_{\text{11}}$ is much more different.

% \section{Algorithm for WID}
% \label{appendix-alo}
% \input{tables/algorithm}

\end{document}